\pdfoutput=1

\documentclass{article}

\usepackage[square,numbers]{natbib}
\usepackage[preprint]{neurips_2025}
\makeatletter
\renewcommand{\@noticestring}{}
\makeatother

\usepackage{times}
\usepackage{latexsym}

\usepackage[T1]{fontenc}

\usepackage[utf8]{inputenc}

\usepackage{microtype}

\usepackage{inconsolata}

\usepackage{graphicx}
\graphicspath{ {./figures/} }
\usepackage{subcaption}

\usepackage{xcolor}
\usepackage{comment}
\usepackage{multirow}
\usepackage{enumitem}
\usepackage{longtable}

\usepackage{hyperref}       
\usepackage{url}            
\usepackage{booktabs}       
\usepackage{amsfonts}       
\usepackage{nicefrac}       

\title{Evaluating the Sensitivity of LLMs to Prior Context}

\author{
  \textbf{Robert Hankache\textsuperscript{1}} \quad
  \textbf{Kingsley Nketia Acheampong\textsuperscript{1}} \quad
  \textbf{Liang Song\textsuperscript{1}}
  \\
  \textbf{Marek Brynda\textsuperscript{1}}\quad
  \textbf{Raad Khraishi\textsuperscript{1,2}} \quad
  \textbf{Greig A. Cowan\textsuperscript{1}}
\\
\\
  \textsuperscript{1}NatWest AI Research \\
  \textsuperscript{2}University College London
}

\begin{document}
\maketitle

\begin{abstract}
\label{sec:abstract}

As large language models (LLMs) are increasingly deployed in multi-turn dialogue and other sustained interactive scenarios, it is essential to understand how extended context affects their performance. 
Popular benchmarks, focusing primarily on single-turn question answering (QA) tasks, fail to capture the effects of multi-turn exchanges. 
To address this gap, we introduce a novel set of benchmarks that systematically vary the volume and nature of prior context. 
We evaluate multiple conventional LLMs, including GPT, Claude, and Gemini, across these benchmarks to measure their sensitivity to contextual variations. 
Our findings reveal that LLM performance on multiple-choice questions can degrade dramatically in multi-turn interactions, with performance drops as large as 73\% for certain models. 
Even highly capable models such as GPT-4o exhibit up to a 32\% decrease in accuracy. 
Notably, the relative performance of larger versus smaller models is not always predictable. 
Moreover, the strategic placement of the task description within the context can substantially mitigate performance drops, improving the accuracy by as much as a factor of 3.5. 
These findings underscore the need for robust strategies to design, evaluate, and mitigate context-related sensitivity in LLMs.

\end{abstract}

\section{Introduction}
\label{sec:intro}

The rapid evolution and deployment of large language models (LLMs) in various natural language processing applications has sparked substantial interest in understanding the capacity of models when handling complex contextual settings and interactions in natural language, particularly, settings that involve conversation sequences. 
With LLMs increasingly underpinning applications across domains such as conversational agents \cite{wang2023enabling, achiam2023gpt}, content generation \cite{moore2023empowering}, document editing \cite{laban2024beyond}, and educational tutoring systems \cite{gan2023large}, there is a pressing need to evaluate their robustness beyond single-turn tasks.

A critical aspect of LLM usage in real-world systems is their ability to manage dynamic contextual information across varied multi-turn user interaction settings \cite{yang2024zhongjing, yi2024survey}. 
In such settings, the sequence of interactions can vary greatly in terms of domain specificity, task type, and semantic coherence.
While LLMs have been successful in tasks such as few-shot learning \cite{perez2021true}, where minimal context is provided to infer or adapt to new tasks, and in addressing needle-in-haystack problems \cite{nelson2024needle}, where the goal is to extract or identify highly specific information from extensive datasets, these tasks fundamentally differ from maintaining contextual coherence across extended multi-turn interactions. 
Recent insights from research have shown that maintaining coherence and accuracy over multiple interactions is a challenging task for LLMs, often leading to degradation in response quality \cite{yu2024cosafe, fan2024fairmt}.
Few-shot learning often relies on pre-curated prompts or demonstrations to guide the model, while needle-in-haystack challenges prioritize locating relevant information in a static context. 
In contrast, multi-turn interactions demand robust memory mechanisms to interpret and effectively integrate prior conversational turns \cite{zhang2025survey}.
It is also important in long-running agentic AI systems that require robustness and consistency \cite{okpala2025agenticaisystemsapplied}.

Evaluating model performance in multi-turn conversations remains a critical challenge.
While existing single question-answer (Single QA) benchmarks such as MMLU and GPQA have provided valuable insights into various aspects of LLM capabilities, they fall short in adequately assessing multi-turn interactions and the subtle role of context within these settings \cite{yi2024survey}. 

In this study, we address the gap in understanding how LLMs cope with extended multi-turn interactions by demonstrating that performance consistency is strongly influenced by the nature of preceding interactions. 
Also, we assess the performance of smaller and larger LLMs under varying lengths of context.
Specifically, we assess whether models maintain domain consistency if introduced to different context and lengths. We introduce new benchmarks derived from GPQA Diamond \cite{rein2023gpqa} to evaluate the sensitivity of LLMs to prior context to investigate these challenges.
We then conduct a series of controlled experiments on popular LLMs, including GPT, Claude and Gemini models. 
Our experiments systematically vary both the knowledge domain of preceding interactions and the depth of conversation history, analysing LLM performance degradation in the presence of prior context.
Our findings suggest that performance may degrade drastically with long prior context, as high as 73\% drop compared to performance when no prior context is added.
This demonstrates the invalidity of using a single-turn benchmark such as GPQA as a proxy for performance over multi-turn interactions.
As such, our new benchmarks enable the assessment of LLMs' multi-turn capabilities and contextual awareness, which are crucial for enhancing LLM performance in practical scenarios. 
Moreover, this study further reinforces the necessity for design improvements in LLMs, such as mechanisms for better contextual memory management and strategies for handling variable coherence in sequential user interactions to enhance the reliability of these models in dynamic in real-world applications.
\section{Related Work}
\label{sec:related_work}

The ability of LLMs to incorporate and leverage prior context has been a focal point of recent work. 
\citet{floridi2020gpt} introduced GPT-3, demonstrating the model's remarkable zero-shot and few-shot learning capabilities while also highlighting its sensitivity to prompt phrasing and context design. 
Subsequent iterations, including OpenAI's GPT-4 and o3, and other open-source models like Deepseek-R1, have sought to improve the robustness and adaptability of the models to understanding context \cite{achiam2023gpt}. 
However, prior studies have shown that LLMs struggle to maintain coherence across long multi-turn conversations, especially when prior context contains minimal information to aid the LLMs in reasoning \cite{wei2022chain, laban2025llms}. 

To better evaluate model performance, previous studies have presented novel benchmarks for context understanding using multiple datasets or through contrastive decoding \cite{zhu2024large, zhao2024enhancing}. 
These LLM benchmarks, which include HellaSwag, BigBench, TruthfulQA, and Chatbot Arena \cite{zellers2019hellaswag, luo2024bigbench, lin-etal-2022-truthfulqa}, have emerged as standards for assessing LLM performance. 
However, LLM benchmarks have certain limitations \cite{mcintosh2024inadequacies}. Issues such as data contamination \cite{xu2024benchmarkingbenchmarkleakagelarge}, narrow focus \cite{mcintosh2024inadequacies}, and an emphasis on single-turn interactions have been previously reported. 
New benchmarks, such as MT-Eval \cite{kwan2024mt}, are being introduced to handle the evaluation of multi-turn and multilingual interactions; for example, Multi-IF utilizes a hybrid framework that combines LLMs and human annotators \cite{he2024multi}.

Moreover, early models, such as GPT-1, were limited to 512 tokens, which constrained their ability to generate coherent responses over extended inputs \cite{radford2019language}. 
Recent advances have dramatically increased these limits, with many models supporting hundreds or even millions of input tokens.
The ability to handle longer context lengths has enabled LLMs to process larger volumes of information, thereby improving performance in tasks such as summarisation and dialogue systems \cite{yates2021pretrained}. 
However, increasing context length introduces challenges such as the "lost in the middle" phenomenon \cite{brown2020language}, and performance degradation, particularly when domains shift mid-conversation \cite{li2024long, zhang2024bench}. 
Several efforts have aimed to address the limitations of large language models (LLMs) in handling context \cite{choromanski2020rethinking, wang2024beyond, ding2024longrope, wu2022stateful, wang2024augmenting, lester2021power, liu2024balancing}, and consequently, necessitating development of novel evaluation benchmarks.



\section{Methodology}
\label{sec:methodology}

In this section, we describe how we create our benchmark datasets for evaluating the QA performance of LLMs in situations with varying past context types, specify the LLMs we evaluated, and the evaluation criteria for assessing models' sensitivity to prior context.

\subsection{Datasets}
\label{sec:methodology - datasets}
We design experiments to mimic a user's textual interactions with an LLM assistant 
or a chatbot. Each experiment is formed from three inputs:

\begin{enumerate}
	\setlength{\itemsep}{0pt}
    \item An optional prior context or conversational history.
    \item A target query ($q_t$).
    \item Four answer choices for each $q_t$, where only one choice
    is the correct response.
\end{enumerate}
To evaluate model performance, we measure the model's accuracy in selecting the correct response.

\subsubsection{Target Queries}
\label{sec:methodology - target - queries}

All target queries, $q_t$, on which the models are evaluated, are sourced from the STEM categories: 
Biology, Physics, and Chemistry of the Graduate-level Professional Question Answering (GPQA) benchmark dataset \cite{rein2023gpqa} used by top AI institutions to evaluate their LLMs' expert-level reasoning in STEM fields. 
In this study, the GPQA questions are presented to LLMs following the prior contexts to evaluate their ability to maintain accuracy and domain fidelity. 
We use the diamond variant of GPQA to align with benchmark consistency in evaluating scientific reasoning and knowledge \cite{achiam2023gpt,gpt4osystemcard,team2024gemini,guo2025deepseek} which consists of 198 graduate-level multiple-choice questions selected to challenge highly capable and motivated PhD-level non-experts. 
Only questions where experts answer correctly, and the majority of non-experts answer incorrectly are in this variant.
Each target query from GPQA, $q_t$, is accompanied by four of answer choices.
For each query instance presented to the experimental models, we shuffle the order of answer 
choices randomly to mitigate positional bias.

\subsubsection{Prior Context}
\label{sec:methodology - prior - context}

Variety of context is crucial to evaluating LLMs' sensitivity in different context scenarios. As such, prior context was sourced from two main data sources: LMSYS-Chat-1M dataset \cite{zheng2024lmsyschat1mlargescalerealworldllm}, and MMLU dataset \cite{hendrycks2020measuring}. Context lengths for the created dataset range from 4k to 64k tokens.\footnote{The context length (tokens count) is computed using the OpenAI tokeniser for GPT-4o and GPT-4o-mini.}\\

\noindent\textbf{Free-chat.}
The \textit{free-chat} context scenario is used to evaluate how authentic, open-ended prior human interactions with LLMs affect their performance. In real-world use, users frequently engage in highly varied and non-linear conversations, often shifting rapidly between lines of thought as seen in LMSYS-Chat-1M dataset.
  
The LMSYS-Chat-1M dataset is a large-scale corpus of approximately one million real-world user chats with LLMs.
It comprises both single- and multi-turn dialogues originally collected from public chatbot conversations.
For our study, we selected chat samples generated by three of the top-performant models in the dataset: vicuna-33b, llama-2-13b-chat, and mpt-30b-chat.
The selection contains enough volume of conversations for a varied experimentation, more than 45k single- and multi-turn chats.
Additionally, as a preprocessing step to maintain data quality, we removed any empty messages from the sampled conversations, eliminating minor artifacts likely introduced by the original LMSYS-Chat-1M cleaning process.\\

\noindent\textbf{Multi-turn QA.} The multi-turn QA context scenarios make use of multiple STEM and Non-STEM examples large enough to cover thousands of tokens in context length.
Given that the GPQA dataset is restricted to STEM subjects and has a limited size, we chose to utilize an alternative dataset for the prior context to enhance the scope of our experiments.
We selected the Measuring Massive Multitask Language Understanding (MMLU) dataset, due to its volume, broad domain coverage, and high-quality multiple-choice format, ensuring that prior context is both structured and relevant for testing.
The data, used to test models multitask accuracy, covers 57 domains including STEM and Non-STEM subjects.
It was designed to be more challenging than then-existing benchmarks such as General Language Understanding Evaluation (GLUE) on which new language models were achieving better-than-human accuracy.
Similar to GPQA, the benchmark is used as an industry standard for analysing performance of LLM models. We employed the MMLU dataset for generating prior context in two experimental settings:

\begin{itemize}
	\item \textbf{Same-domain Context.} For this setting, we extract STEM-related question-response $qr$ 
	from the MMLU dataset. These include subjects such as Biology, Physics, Chemistry, Clinical Knowledge, 
	and Mathematics.
	\item \textbf{Cross-domain Context.} For this setting, Non-STEM $qr$ pairs from the MMLU dataset are selected. These include
	History, Law, Social Sciences, and Philosophy subjects.
\end{itemize}
See Appendix \ref{sec:appendix - mmlu} for a complete list of selected subjects.

\subsection{Task Position}
\label{sec:methodology - task position}
The task description\footnote{Our prompts follow OpenAI "simple-evals" framework.} for the target query, which includes the expected answer formatting, is placed in the last message after the prior context and before the target query for the \textit{free-chat} context, since the tasks are not related.
For the multi-turn QA experiments, we evaluate the following two settings.
First is \textit{task-at-top}, where the task description is provided only once at the very top of the prior context, i.e. the first message of the conversation.
This priming technique is motivated in \citet{brown2020language} and \citet{liu2023pre}.
Given the initial low performances (see section \ref{sec:results - analysis}) and as a result of the detailed investigations (see Section \ref{sec:results - task - location}), we also evaluate a second setting, \textit{task-repeated}, where the task description is reiterated both at the very top of the prior context, and right before the target query.

\subsection{Models}
\label{sec:methodology - models}
We evaluate several state-of-the-art large language models, namely:
GPT-4o-mini, GPT-4o \cite{gpt4osystemcard}, 
Claude Haiku, Claude Sonnet\footnote{Claude models only accept a maximum of one thousand conversational messages. In cases where we exceed this limit, we combine several rounds of conversation into one message, respecting the conversation formatting and keeping the same content.},
Gemini Flash, and Gemini Pro \cite{team2024gemini} (refer to Appendices \ref{appendix:models_versions} and \ref{appendix:experiments_cost} for models versions and experimental cost). 
These LLMs were selected purposefully for their demonstrated ability in multi-turn conversations and wide use in the AI community, also accounting for the distinct subtleties inherent in the architectures and training paradigms of various providers, thereby ensuring the generalisability of our findings across various LLM implementations.

\noindent\textbf{Temperature Settings.} Various temperature settings, varying from 0 to 1, have been used when 
evaluating LLM models on Question Answering, Maths and reasoning tasks. For our experiments, a temperature 
setting of 0.5 is used for all models to balance creativity, randomness, and determinism in responses \cite{achiam2023gpt, gpt4osystemcard, guo2025deepseek}. 
Accordingly, the models' seeds are varied across evaluation runs to assess 
the models' response variability.



\section{Results}
\label{sec:results}

\begin{table*}[h!]
    \centering
    \caption{LLM performance results, in terms of accuracy, across different context scenarios (various sources and lengths). Accuracies are reported as mean values of three runs, with standard deviations. Context length refers to token counts of the entire text sent to the LLM, which includes the target query and the prior context if added. "No context" refers to the scenario in which no additional context, apart from the target query, is provided to the models.
    }
    \resizebox{\textwidth}{!}{%
    \begin{tabular}{|c|c|c|c|c|c|c|c}
    \hline

\multirow{2}{*}{\textbf{Model}}	
& \multirow{2}{*}{\textbf{Context Length}}		
& \multirow{2}{*}{\textbf{Free-chat}}		
& \multicolumn{2}{c|}{\textbf{Same Domain}}     		
& \multicolumn{2}{c|}{\textbf{Cross Domain}}  \\
\cline{4-7}
    				
    			&          		&          	
& \multicolumn{1}{c|}{\textbf{Task-at-top}} 
& \textbf{Task-repeated} 		
& \multicolumn{1}{c|}{\textbf{Task-at-top}} 	
& \textbf{Task-repeated} \\ \hline
    
    \multirow{4}{*}{Claude Haiku}    & no context 	& 0.365$\pm0.020$ 				& 0.365$\pm0.020$ 				& 0.365$\pm0.020$ 				& 0.365$\pm0.020$ 				& 0.365$\pm0.020$ \\
                                     & 4k  			& 0.365$\pm0.021$ 				& 0.357$\pm0.018$ 				& 0.362$\pm0.029$ 				& 0.343$\pm0.013$ 				& 0.386$\pm0.023$ \\
                                     & 16k 			& 0.332$\pm0.036$ 				& 0.360$\pm0.015$ 				& 0.318$\pm0.035$ 				& 0.323$\pm0.013$ 				& 0.367$\pm0.023$ \\
                                     & 32k 			& 0.354$\pm0.036$ 				& 0.359$\pm0.005$ 				& 0.372$\pm0.018$ 				& 0.350$\pm0.008$ 				& 0.367$\pm0.034$ \\
                                     & 64k 			& 0.352$\pm0.006$ 				& 0.347$\pm0.031$ 				& 0.359$\pm0.005$ 				& 0.315$\pm0.006$ 				& 0.345$\pm0.042$ \\ \hline
    
    \multirow{4}{*}{Claude Sonnet} 	& no context   	& 0.412$\pm0.008$ 				& 0.412$\pm0.008$				& 0.412$\pm0.008$				& 0.412$\pm0.008$ 				& 0.412$\pm0.008$ \\
                                     & 4k  			& 0.375$\pm0.034$ 				& 0.375$\pm0.043$ 				& 0.389$\pm0.009$ 				& 0.249$\pm0.034$ 				& 0.386$\pm0.006$ \\
                                     & 16k 			& 0.367$\pm0.019$ 				& 0.375$\pm0.008$ 				& 0.411$\pm0.028$ 				& 0.172$\pm0.017$ 				& 0.407$\pm0.028$ \\
                                     & 32k 			& 0.382$\pm0.011$ 				& 0.350$\pm0.030$ 				& 0.389$\pm0.010$ 				& 0.180$\pm0.037$ 				& 0.412$\pm0.023$ \\
                                     & 64k 			& 0.384$\pm0.031$ 				& 0.380$\pm0.016$ 				& 0.375$\pm0.025$ 				& 0.190$\pm0.051$ 				& 0.362$\pm0.044$ \\ \hline
    
    \multirow{4}{*}{Gemini Flash}	& no context   	& 0.476$\pm0.006$ 				& 0.476$\pm0.006$				& 0.476$\pm0.006$ 				& 0.476$\pm0.006$ 				& 0.476$\pm0.006$ \\
                                     & 4k  			& 0.412$\pm0.008$ 				& 0.313$\pm0.040$ 				& 0.470$\pm0.025$				& 0.305$\pm0.023$ 				& 0.446$\pm0.006$ \\
                                     & 16k 			& 0.421$\pm0.011$ 				& 0.340$\pm0.008$ 				& 0.480$\pm0.018$ 				& 0.180$\pm0.015$ 				& 0.443$\pm0.029$ \\
                                     & 32k 			& 0.439$\pm0.020$ 				& 0.354$\pm0.005$ 				& 0.461$\pm0.020$ 				& 0.162$\pm0.027$ 				& 0.443$\pm0.008$ \\
                                     & 64k 			& 0.421$\pm0.034$ 				& 0.340$\pm0.028$ 				& 0.475$\pm0.015$ 				& 0.128$\pm0.026$ 				& 0.438$\pm0.021$ \\ \hline
    
    \multirow{4}{*}{Gemini Pro}		& no context   	& 0.545$\pm0.013$ 				& 0.545$\pm0.013$				& 0.545$\pm0.013$ 				& 0.545$\pm0.013$ 				& 0.545$\pm0.013$ \\
                                     & 4k  			& 0.545$\pm0.018$ 				& 0.483$\pm0.018$ 				& 0.537$\pm0.016$ 				& 0.502$\pm0.008$ 				& 0.517$\pm0.011$ \\
                                     & 16k 			& 0.488$\pm0.015$ 				& 0.480$\pm0.033$ 				& 0.539$\pm0.011$ 				& 0.471$\pm0.008$ 				& 0.551$\pm0.033$ \\
                                     & 32k 			& 0.502$\pm0.030$ 				& 0.500$\pm0.033$ 				& 0.552$\pm0.018$ 				& 0.451$\pm0.008$ 				& 0.502$\pm0.013$ \\
                                     & 64k 			& 0.481$\pm0.044$ 				& 0.473$\pm0.013$ 				& 0.552$\pm0.041$ 				& 0.460$\pm0.028$ 				& 0.534$\pm0.013$ \\ \hline
    
    \multirow{4}{*}{GPT-4o}		  	& no context   	& 0.524$\pm0.019$ 				& 0.524$\pm0.019$				& 0.524$\pm0.019$				& 0.524$\pm0.019$ 				& 0.524$\pm0.019$ \\
                                     & 4k  			& 0.453$\pm0.030$ 				& 0.449$\pm0.023$ 				& 0.500$\pm0.015$ 				& 0.461$\pm0.046$ 				& 0.510$\pm0.018$ \\
                                     & 16k 			& 0.461$\pm0.031$ 				& 0.439$\pm0.010$ 				& 0.510$\pm0.022$ 				& 0.389$\pm0.018$ 				& 0.502$\pm0.012$ \\
                                     & 32k 			& 0.451$\pm0.013$ 				& 0.423$\pm0.016$ 				& 0.475$\pm0.013$ 				& 0.372$\pm0.015$				& 0.525$\pm0.015$ \\
                                     & 64k 			& 0.458$\pm0.013$ 				& 0.441$\pm0.008$ 				& 0.483$\pm0.016$ 				& 0.355$\pm0.024$ 				& 0.510$\pm0.028$ \\ \hline
    
    \multirow{4}{*}{GPT-4o-mini} 	& no context   	& 0.451$\pm0.044$ 				& 0.451$\pm0.044$				& 0.451$\pm0.044$				& 0.451$\pm0.044$ 				& 0.451$\pm0.044$ \\
                                     & 4k  			& 0.412$\pm0.011$ 				& 0.348$\pm0.036$ 				& 0.394$\pm0.013$ 				& 0.320$\pm0.016$ 				& 0.423$\pm0.019$ \\
                                     & 16k 			& 0.406$\pm0.026$ 				& 0.276$\pm0.013$ 				& 0.409 $\pm0.026$ 				& 0.226$\pm0.024$ 				& 0.423$\pm0.018$ \\
                                     & 32k 			& 0.396$\pm0.026$ 				& 0.278$\pm0.028$ 				& 0.382$\pm0.023$ 				& 0.190$\pm0.006$ 				& 0.418$\pm0.011$ \\
                                     & 64k 			& 0.392$\pm0.048$ 				& 0.293$\pm0.022$ 				& 0.404$\pm0.045$ 				& 0.205$\pm0.028$ 				& 0.416$\pm0.025$ \\ \hline
    \end{tabular}%
    }

    \label{table:1}
    \end{table*}

Using the context scenarios described in the previous section, we run several experiments to evaluate the sensitivity of the selected large language models to prior context.
We set up two primary experimental conditions.
In each, the target query is a STEM-related query from the GPQA dataset (Section \ref{sec:methodology - target - queries}).
In the first experiment, prior context consists of multi-turn \textit{free-chat} conversations (Section \ref{sec:methodology - prior - context}).
The task description and target query ($q_t$) are added as the last message in the sequence.
In the second experiment, we present each model with prior context consisting of \textit{same-} or \textit{cross-domain} multi-turn QA conversation-styled interactions (Section \ref{sec:methodology - prior - context}).
In addition, the location of the task description in the prior context is varied between \textit{task-at-top} and \textit{task-repeated} scenarios, to better ascertain task location influence on the overall model performance.
Hence in this second experiment, the total unique conditions are four.
In all experiments, the various LLMs were given the same list of target queries and additional context. 
For every target query, $q_t$, in the experiments, we present models with four varying lengths of prior context: 4k, 16k, 32k and 64k. 
The variation in context lengths enables us to not only assess sensitivity in the presence of context, but also the effect of increasing prior context lengths. 

Only complete $qr$ pairs are presented as input context; a soft truncation is applied on the $qr$ level.
We add a \textit{no-context}  scenario which serves as a base experiment during the context sensitivity experiments.\footnote{For the base \textit{no-context}  experiments, the average context length of the target queries is 270 tokens.} 
Three evaluation runs across three different seeds are used for each combination of queries, context types, and models.
The following randomisations are done for each run: i) shuffle the target query's choices, ii) randomise the additional prior context, and iii) change the model's seed. 
The mean and standard deviation of the accuracy is computed from the three runs and are shown in Table \ref{table:1}.
In addition to absolute values, relative scores with respect to the base \textit{no-context}  experiments are recorded in the appendix (see Table \ref{appendix: results - full}). We now discuss the results of the \textit{free-chat} experiments and multi-turn QA experiments.


\subsection{Free-chat Context Experiments}
\label{sec:results - free-chat}

\begin{figure}[t]
    \centering
    \includegraphics[width=0.49\textwidth]{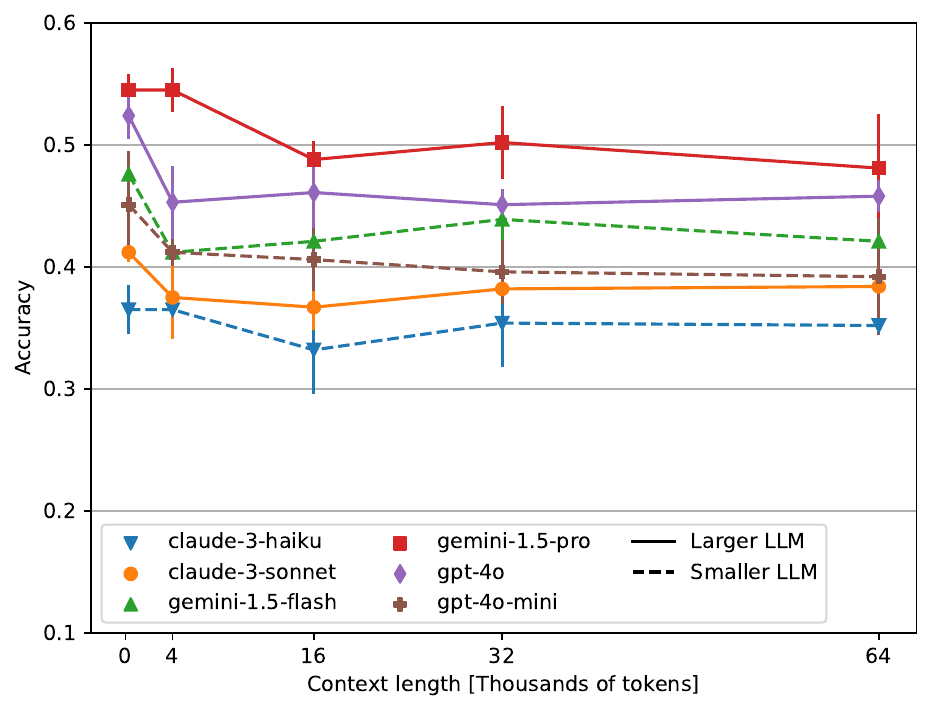}
    \caption{Experimental LLM performance over variable \textit{free-chat} context. Scores are mean accuracies across three runs with the standard deviations shown as error bars. Dashed and continuous lines represent smaller and larger LLMs respectively.
}
    \label{fig:free_chat}
\end{figure}

\noindent\textbf{Prior context leads to performance degradation.}
It is clear from Figure \ref{fig:free_chat} that prior contexts cause models' performances to degrade.
This is true for all the six models we tested from the three different providers.
It is reasonable to assume that the inclusion of unrelated free chat would not influence the model's performance, particularly since the target question is independent of information from the prior context.
Nonetheless, the performance drops observed are non-negligible.
The average accuracy drops in the highest two bins ranges from 3.5\% to 13\% relative to \textit{no-context}  scenarios, revealing a varied effect depending on the model, and especially between model providers.
Although the degradation is bigger for some models, we do not observe any flip in accuracy rankings.
Instead, we see a reduction of the accuracy spread within the same prior context experiments.
With \textit{no-context} , the top performing model (Gemini Pro) accuracy is 49\% higher than the worst performing model (Claude Haiku).
At 64k-tokens prior context, this difference is reduces to 37\%.

\noindent\textbf{Model performance degradation is affected by the length of interactions.}
The length of multi-turn interactions also plays a critical role in performance stability, as expected. 
The longer the conversational interaction, the higher the drop in performance that is observed for all the models, up to a certain length.
After 4k to 16k context length, models' performances drop very slightly or even stabilise.
For instance, for the top performing model Gemini Pro, the accuracy drops 5.7\% from no- to 16k-token context; then the remaining drop until 64k-token context is just 0.7\%.
At 4k-token context, GPT-4o model had the largest relative drop of 12\%, with a negligible remaining drop until 64k-token context.
This early stabilisation might be due to the following: queries $q_t$, that were correctly answered in the \textit{no-context}  scenario but with lower confidence, might have their answers easily flipped even with short prior context; whereas the ones answered with high confidence will have the same answer even at 64k context lengths.
More studies are needed to understand and explain this behaviour.

\noindent\textbf{Larger models do not necessarily improve robustness to context sensitivity.}
Contrary to our expectations, we found large models to be just as susceptible to the effect of prior context as their smaller counterparts.
Comparing models within each family, smaller and larger GPT (Gemini) models have similar accuracy drops in the highest two bins, averaging 13\% (12\%) (see Figure \ref{fig:free_chat}).
Claude models exhibit a different effect; in those same two bins, the larger Sonnet model's performance drops 7\%, compared to 3.5\% only for the smaller Haiku model.
The additional intriguing observation is that Claude Haiku, which is the least affected by the prior context, is at the same time the worst performing model between the six.
This shows empirically that it is not guaranteed that a more powerful and high-performing model in basic benchmarks will maintain its advantage in other scenarios with varying context.


\subsection{Multi-turn QA Context Experiments}
\label{sec:results - analysis}
\label{sec:results - qa}

\begin{figure}[tb] 
    \begin{subfigure}[b]{0.45\textwidth}
        \centering
        \includegraphics[width=\textwidth]{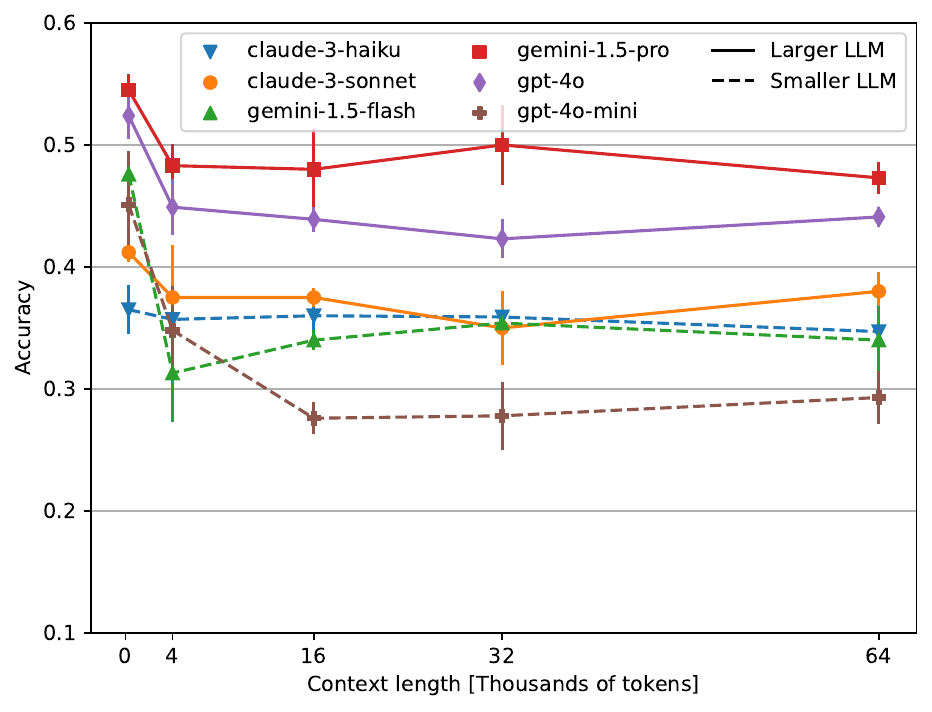}
    \caption{QA \textit{task-at-top} \textit{same-domain} context}
    \label{fig:same_domain_task_at_top}
    \end{subfigure}
    \hfill
    \begin{subfigure}[b]{0.45\textwidth}
        \centering
        \includegraphics[width=\textwidth]{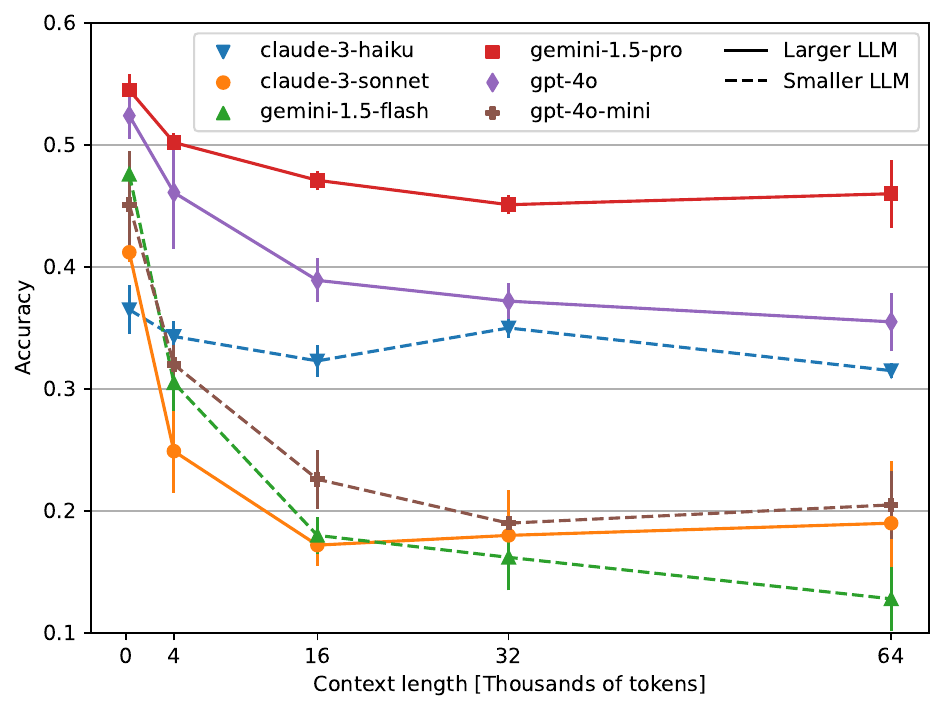}
    \caption{QA \textit{task-at-top} \textit{cross-domain} context}
    \label{fig:cross_domain_task_at_top}
    \end{subfigure}

    \begin{subfigure}[b]{0.45\textwidth}
        \centering
        \includegraphics[width=\textwidth]{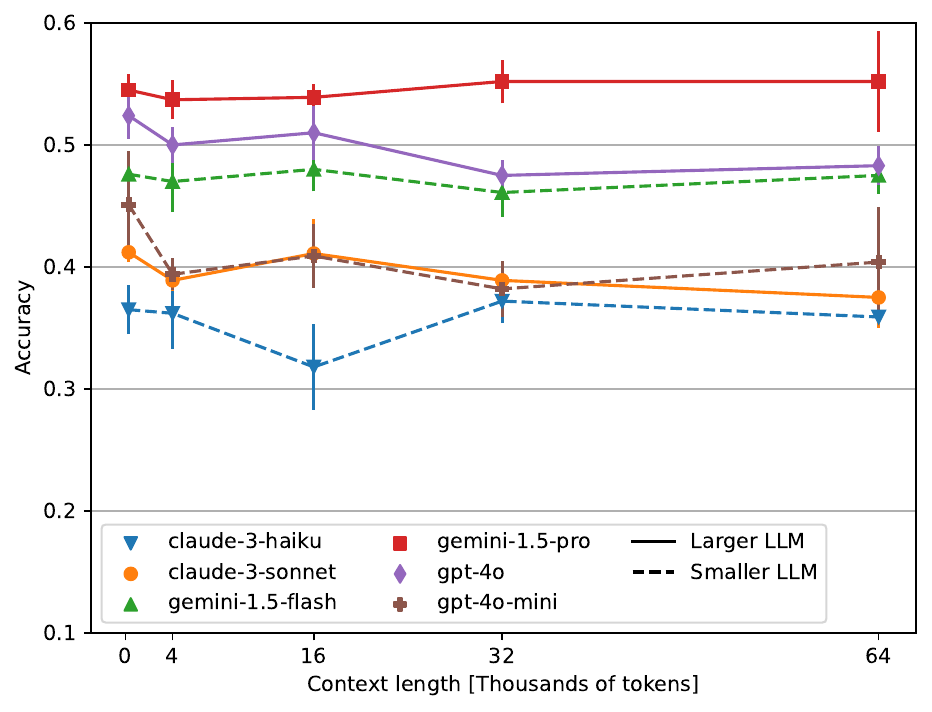}
    \caption{QA \textit{task-repeated} \textit{same-domain} context}
    \label{fig:same_domain_task_repeated}
    \end{subfigure}
    \hfill
    \begin{subfigure}[b]{0.45\textwidth}
        \centering
        \includegraphics[width=\textwidth]{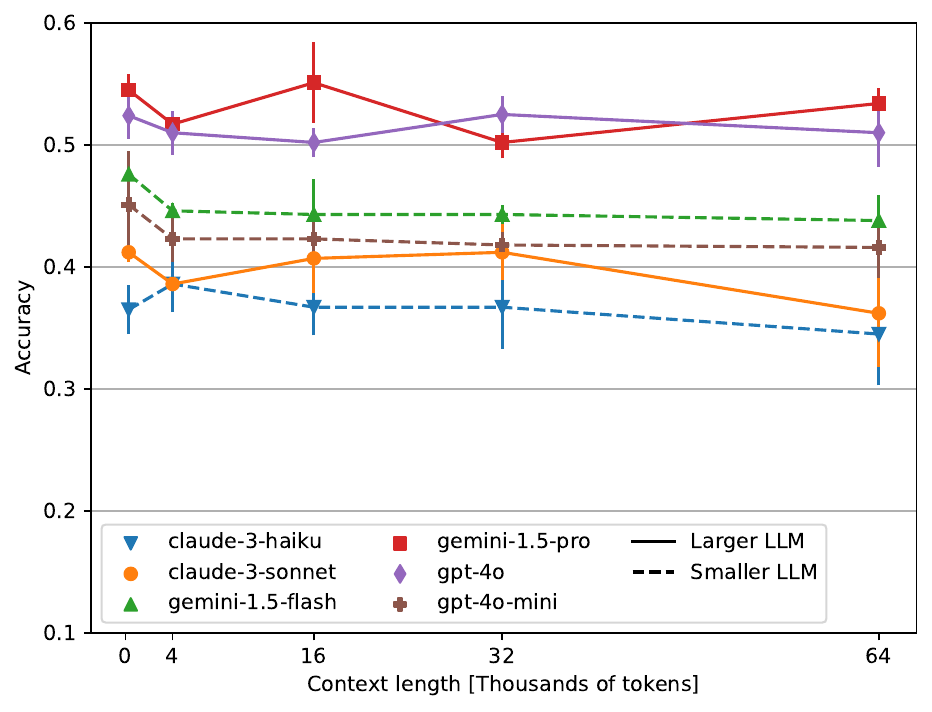}
    \caption{QA \textit{task-repeated} \textit{cross-domain} context}
    \label{fig:cross_domain_task_repeated}
    \end{subfigure}

    \caption{Experimental LLM performance over variable multi-turn \textit{task-at-top} (top row), \textit{task-repeated} (bottom row), \textit{same-domain} (left column) and \textit{cross-domain} (right column) contexts. Scores are mean relative accuracies across three runs with the standard deviations shown as error bars. Dashed and continuous lines represent smaller and larger LLMs respectively.}
    \label{fig:qa_results}
\end{figure}

\noindent\textbf{Prior interactions have large impact on models' performances.}
In Figures \ref{fig:same_domain_task_at_top} and \ref{fig:cross_domain_task_at_top}, we see the effect of multi-turn QA context on model performance. Notably, we find that the effect of context is more significant for the \textit{task-at-top} scenario, where most models suffer from significant performance degradation, with some experiencing as high as a 73\% relative drop (Gemini Flash with \textit{cross-domain} context).
Although locating the task at the top of the context is intuitive and well-motivated as the task does not change with our final query similar to few-shot prompting scenarios, we see that having any type of context preceding a target query, whether it's QA from the \textit{same-} or \textit{cross-domain}, significantly influences the models' performances. 
The only exception observed among the six models is Claude Haiku, which remains almost flat for the \textit{same-domain} experiments.
Furthermore, it is interesting to see the models' ranking change between no prior context and the longest context scenario.
In \textit{same-domain} experiments, both Claude models climbed two ranks up from the bottom.
In \textit{cross-domain} experiments, Claude Haiku went up from last to third place, in contrast with Gemini Flash that went down from third to last place.
The accuracy spread of models increases with the addition of more prior context.
With no prior context, the highest-accuracy over lowest-accuracy ratio is 1.5, but at 64k-tokens it went up to 1.6 for \textit{same-domain} and a staggering 3.6 for \textit{cross-domain} scenarios.

In contrast, the \textit{task-repeated} scenario seems to be much more robust, as can be seen in Figures \ref{fig:same_domain_task_repeated} and \ref{fig:cross_domain_task_repeated}, with a maximum drop for models being 15\% (GPT-4o-mini) and 12\% (Claude Sonnet) for the \textit{same-domain} and \textit{cross-domain} experiments, respectively.
The performance drops for the \textit{task-repeated} scenario are more comparable to the \textit{free-chat} scenario; the only noticeable difference is that Gemini models performed slightly worse in the latter.
Same as with the \textit{free-chat} results, no models ranking changes were observed for the \textit{task-repeated} scenario.
As for the highest-accuracy over lowest-accuracy ratio, it stays almost unchanged for both \textit{same-} and \textit{cross-domain} scenarios between no and longest prior context. This finding motivates repeated task placement as a mitigation strategy for degradation of model performance due to prior context.

\begin{figure}[tb] 
    \centering
    \begin{subfigure}[b]{0.49\linewidth}
        \centering
            \includegraphics[width=\linewidth]{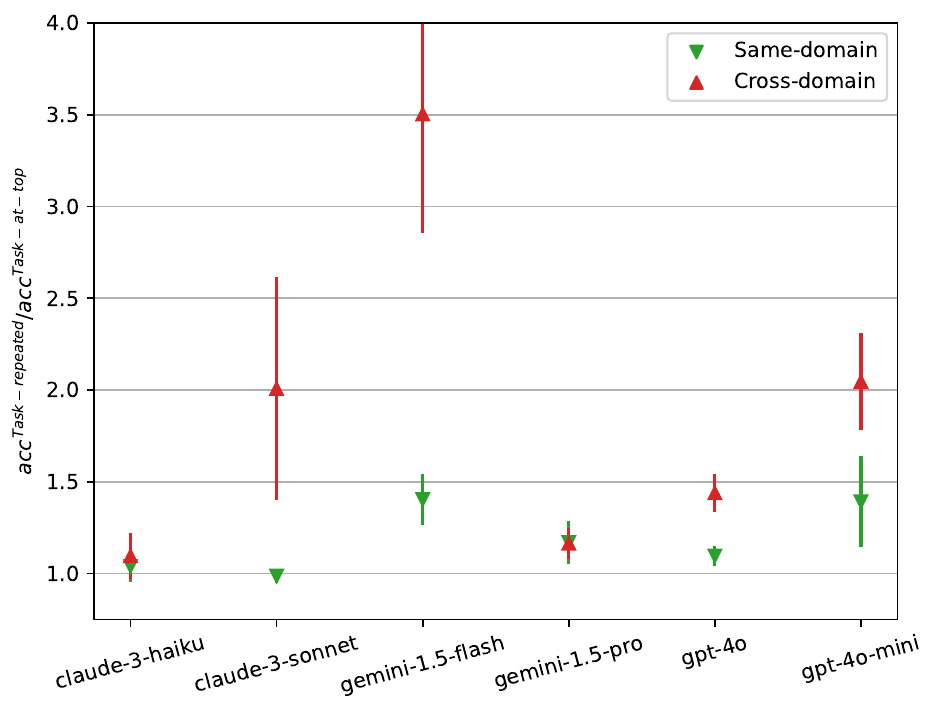}
    \caption{Accuracy ratio}
    \label{fig:acc_ratio_64k}
    \end{subfigure}
    \hfill
    \begin{subfigure}[b]{0.49\linewidth}
        \centering
        \includegraphics[width=\linewidth]{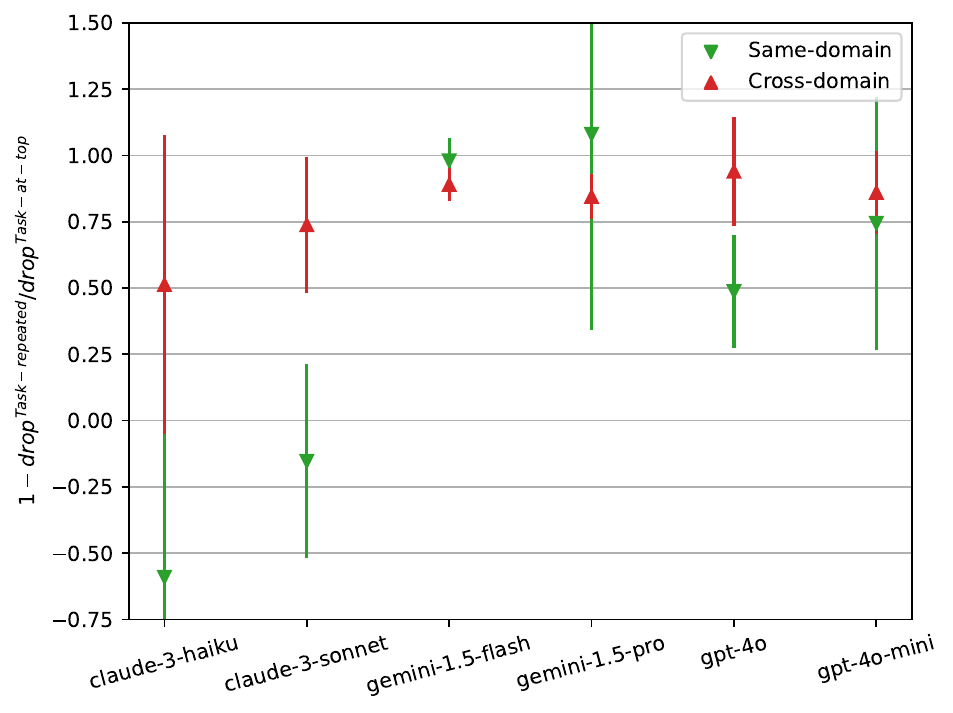}
    \caption{Recovered drop}
    \label{fig:recovered_drop_64k}
    \end{subfigure}
    \caption{Performance improvements when repeating the task description before the query question.
    (a) Accuracy ratio of \textit{task-repeated} over \textit{task-at-top} scenarios for both \textit{same-domain} and \textit{cross-domain} multi-turn QA context and 64k-token length.
    (b) Amount of recovered performance drop after repeating the task (zero means no performance recovered, one mean full performance recovered, values smaller than zero or larger than one are due to statistical fluctuations).}
    \label{fig:task_at_top}
\end{figure}

\noindent\textbf{Task location influences significantly the model performance.}
As highlighted earlier, we find repeated task placement to be an effective mitigation for the performance degrading effects of the prior context.
The accuracies for the \textit{task-repeated} scenario were significantly higher in some experiments compared to \textit{task-at-top} ones.
To better quantify the improvement after repeating the task before the query question, we plot in Figure \ref{fig:acc_ratio_64k} the accuracy ratio of \textit{task-repeated} over \textit{task-at-top} scenarios for both \textit{same-domain} and \textit{cross-domain} context.
The biggest improvements can be seen in \textit{cross-domain} experiments, where Gemini Flash accuracy improved by a factor of 3.4.
For Claude Sonnet, the larger model within the Claude family, the accuracy increased from 19\% to 36.2\% when the task description was repeated in the \textit{cross-domain} context of 64k length, representing nearly a doubling of performance. 
Another way to assess this improvement is to evaluate the amount of performance drop that was recovered after repeating the task description, as illustrated in Figure \ref{fig:recovered_drop_64k}.
For GPT and Gemini models in a \textit{cross-domain} scenario, 85\% or more of the drop is recovered. 
For example, for GPT-4o-mini, the average accuracy drop in the highest two bins is 56\% relative to the base \textit{no-context}  experiment for the \textit{cross-domain} \textit{task-at-top} context.
In the \textit{task-repeated} experiment, this drop is only 8\%, indicating that 86\% of the drop is recovered.
Claude Haiku, which was the most stable in \textit{task-at-top} experiments, has seen the lowest average improvements across the two domains, where both are compatible with no change in accuracy.
This difference in performance regarding the two task-location scenarios is unexpected but has very important implications: shedding light on how LLMs use information in different locations in the context and, as a consequence, on how to structure the prompt for best performance.
These results suggest that reiterating the task description is an effective mitigation technique where long context separates the description and the query.
A detailed investigation of this observation is discussed in Section \ref{sec:results - task - location}.

\noindent\textbf{Larger models performance can become equal to or worse than smaller models.}
Claude Sonnet model, which is larger than Claude Haiku model, has higher performance in the base \textit{no-context}  experiment (+13\% relatively).
For the \textit{same-domain} context in \textit{task-at-top} experiments (Figure \ref{fig:same_domain_task_at_top}), Claude Sonnet's accuracy decreases with more added prior context, until it becomes similar to Claude Haiku's accuracy. 
However, for the \textit{cross-domain} context (Figure \ref{fig:cross_domain_task_at_top}), Claude Sonnet's accuracy drops below that of Claude Haiku after just a few thousand tokens of additional context, reaching as low as half of Haiku's accuracy.
Same as seen in the \textit{free-chat} scenario, this observation shows the importance of benchmarks assessing the effect of prior context.

\noindent\textbf{Model performance worsens when the QA prior context and the target query are from different knowledge domains.}
It can be argued that LLMs may focus on important and relevant information to answer the target query while disregarding non-relevant information. 
However, we found a surprising performance disparity between the \textit{same-domain} and \textit{cross-domain} experiments in the \textit{task-at-top} scenario, as demonstrated empirically through the experiments.
All models performed worse in the \textit{task-at-top} experiments where prior context and target queries came from cross domains (Figure \ref{fig:cross_domain_task_at_top}), as opposed to those from the \textit{same-domain} context (Figure \ref{fig:same_domain_task_at_top}).
For some results at 4k tokens length, the difference is negligible and can even go in the opposite direction due to fluctuations.
In comparison, the models most affected by the \textit{same-domain} context were GPT-4o-mini and Gemini Flash, with average accuracy drops in the highest two bins of 36\% and 27\% relative to the baseline \textit{no-context}  experiment. 
For the \textit{cross-domain} context, however, the average drops in those same bins were 56\% and 70\%.

For the more robust \textit{task-repeated} experiments, the performance differences are significantly smaller between \textit{same-} and \textit{cross-domain} contexts.
The trend of lower performances for the latter is usually observed, except for GPT models, where the accuracy fluctuations are greater than the differences, making it difficult to draw a conclusive observation.

Overall, the observed difference in performance between \textit{same-} and \textit{cross-domain} contexts suggests that the model's internal representations favour context continuity. 
Furthermore, the performance of the models in both \textit{same-domain} and \textit{cross-domain} experiments becomes comparable when the task is repeated, indicating that the context, and the task location within it, significantly influences the contextual behaviour of models more than previously thought.

\noindent\textbf{Model performance declines with lengthier interactions.}
For \textit{same-domain} experiments in the \textit{task-at-top} scenario, the performance drop flattens after 4k tokens, except for the GPT-4o-mini model, which continues to drop until 16k. 
In \textit{cross-domain} experiments, the general trend is that performances keep dropping but at slower rates (except for a few statistically insignificant fluctuations). 
The two most performant models across the various context lengths are GPT-4o and Gemini Pro. Their accuracies drop by 16\% (32\%) and 14\% (16\%) relatively between \textit{no-context}  and 64k-tokens \textit{same-domain} (\textit{cross-domain}) context experiments.

In \textit{task-repeated} scenarios, the model performance variations are small and comparable in magnitude to fluctuations, which makes it harder to detect the general trend in performance.
For some models, most of the degradation occurs in the experiments with a 4k-token context length, while for others, small degradation continues until the 64k-token experiments.
Ultimately, it is important to note that repeating task descriptions within the context proves effective in reducing the repercussions of long context.

\subsubsection{Understanding the Effect of Task Location}
\label{sec:results - task - location}

Given the large performance differences between \textit{task-at-top} and \textit{task-repeated} scenarios, it is important to understand why the task location has this significant influence on models.
The hypothesis is that the task information is being diluted with longer context, and this was the initial motivation for having the two task location scenarios.

Qualitative analysis of the models responses reveals three primary sources of errors contributing to low performance scores: i) reasoning errors, where the model selects the wrong choice; ii) answer format deviations, where the correct choice was selected but the wrong formatting leads to a failed answer extraction; and iii) models not answering the given task, where the output is not related to answering the question (e.g. becoming stuck in repetitive token sampling).

The most significant factor identified for the performance degradation under the \textit{task-at-top} condition was formatting issues, where models frequently diverged from the expected response format and occasionally produced output in unintended formats, such as LaTeX, which contradicted the task description. 
This observation is non-intuitive, since having the task at the beginning primes the models, and all QA examples within the context adhere to the expected answer format, effectively functioning as few-shot learning instances for the models.
While we expected that the adherence to the correct formatting will be reinforced more in such conditions, the results show the opposite effect.

Low model performance scores in \textit{task-at-top} scenarios stems from multiple factors, especially the dilution of task-relevant information in longer contexts. 
Repeating the task towards the end appears to refocus the model's attention on the key instructions, thereby ensuring the output complies with the specified format.
Ultimately, these experimental results emphasize the importance of strategic task placement within the context, particularly for tasks that involve extensive inputs.
By reintroducing task-specific details at key locations, we can greatly improve the performance of language models.
These findings are important for any LLM application where descriptions and guidelines are presented at the beginning of the inputs.
For example, it is common to define the persona, tone of voice, personalisation, and guidelines of chat models in the first conversation message as a system prompt.
Also, in agentic application where multi-round of (self-)interaction are present, the agent guidelines and tools are also defined at the beginning.
In those cases, it is important to evaluate the model on lengthy interactions and anticipate any performance degradation.

\section{Conclusion and Future Work}
\label{sec:conclusion}

In this study, we introduce novel benchmarks comprising various multi-turn conversations, and explore the sensitivity of several LLMs to prior context, evaluating their accuracies and robustness in various contexts and prompting scenarios.
Our findings reveal the substantial influence that prior context as well as priming techniques have on LLM performance, which is more significant than previously thought.
Relative accuracy drops as high as 73\% are observed on GPQA questions.
This finding was unanticipated, especially in some of our scenarios that are similar to few-shots prompting where one would expect more aligned responses.
Our results show the inadequacy of relying on a single-turn benchmark to accurately reflect performance across multi-turn interactions.
The impact of prior context varies depending on the content, even when it is unrelated to the current task.
We also find that multi-turn \textit{cross-domain} QA context harms models' performances even more compared to \textit{same-domain} context.
Moreover, larger models are not necessarily more robust to the effects of prior context, where, in some scenarios, their accuracies dropped below that of smaller models.
Task information dilution contributes significantly to the degradation of accuracy.
Repeating the task before the last query proves to be an effective mitigation strategy.
This approach helps recover a substantial portion of the drop, with many models gaining twice as much or more in accuracy.

This variability highlights the need for a deeper understanding of LLM behaviour in contexts where sustained engagement and contextual awareness are required.
It also emphasises the importance of benchmarks that evaluate the effect of prior context, such as the ones we introduced in this paper. 
This will guide the selection of LLMs that are not only proficient in focused reasoning, but also capable of maintaining consistent performance in real-world applications where context is nearly inevitable.
Moreover, when engineering model prompts, one should account for the risk of information dilution when task description is introduced early in the conversation.

Future research may extend this study to include more varied context, such as different languages and code context, to capture a wider range of real-life scenarios; as well as apply this study to alternative benchmarks other than GPQA. 
Next, we aim to develop methods that assess the relevance of prior 
interactions in context and assess LLMs ability to differentiate between contextually dependent and independent queries, leading to more accurate and coherent responses.
We will also evaluate the effect of chain-of-thought, prior context summarisation and selective memory retention techniques on the sensitivity to prior context.

\section{Limitations}
\label{sec:limitations}

Our analysis is limited by the set of questions in LMSYS-1M-Chat and QA benchmark datasets (i.e., MMLU and GPQA), which may not capture the full diversity prior context LLM users may experience in real-world applications.
In the prior context, the same given answers are used for all models to maintain experimental consistency. Alternatively, one could use the actual model answers to test for additional error propagation.
Moreover, the models were evaluated in controlled settings, which, while necessary for consistency, may not fully represent the variability found in everyday natural language interactions, such as cultural and linguistic diversities, and more varied conversation history.

\section*{Acknowledgements}
\label{sec:acknowledgements}

We wish to express appreciation Graham Smith and Zachery Anderson of NatWest Group for the time and support needed to develop this research paper.

\bibliography{entire}

\newpage
\appendix

\newpage

\section{Illustration}
\begin{figure}[t!]
    \centering
    \includegraphics[width=0.49\textwidth]{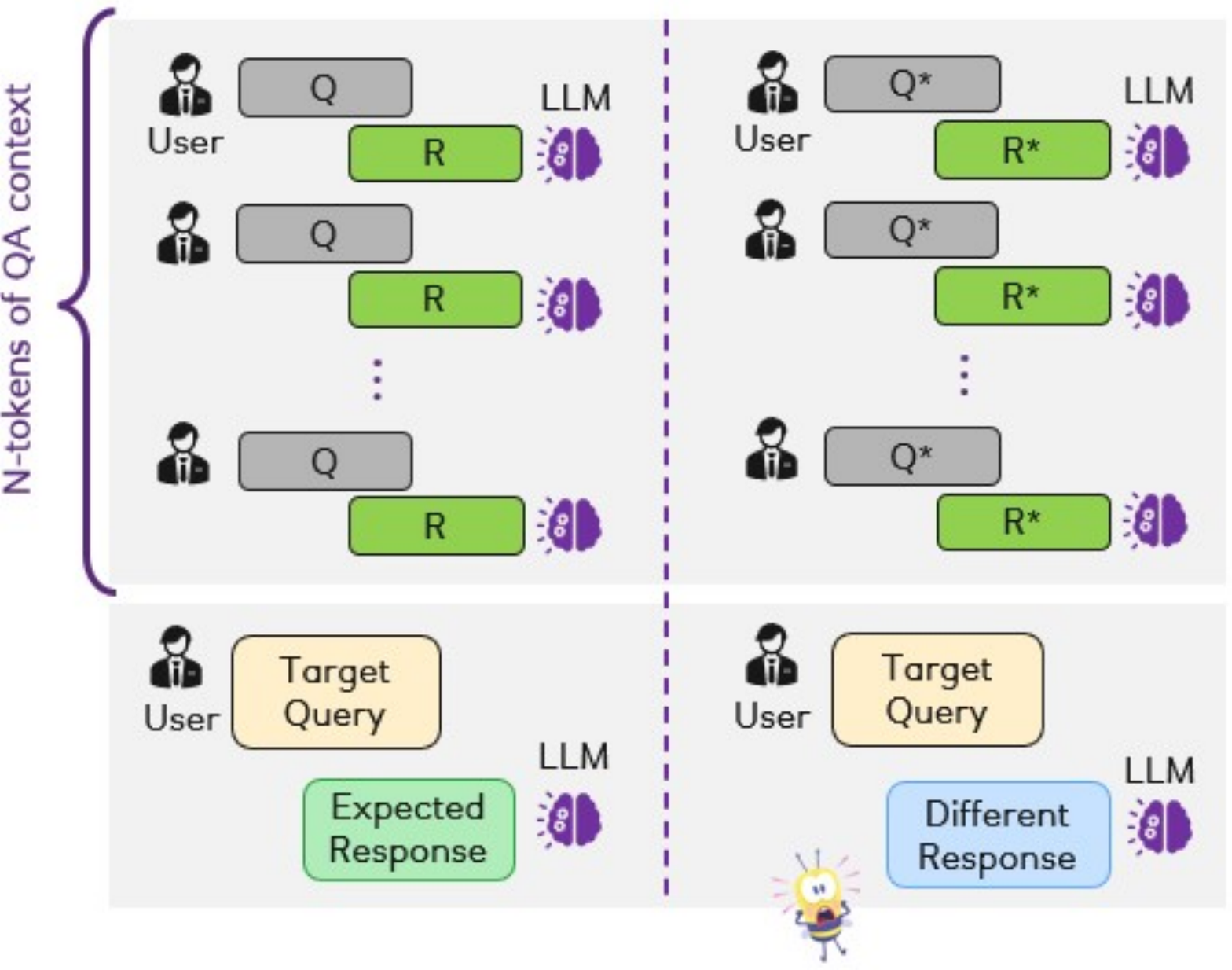}
    \caption{An illustration of varying context scenario in interactive settings where LLMs respond differently to the same target query when preceded with multiple turns of interactions.}
    \label{fig:figure_1}
\end{figure}

Figure \ref{fig:figure_1} illustrates the prior context experimentation idea.

\section{Relative Accuracy Plots}
\begin{figure}[t]
    \centering
    \includegraphics[width=0.49\textwidth]{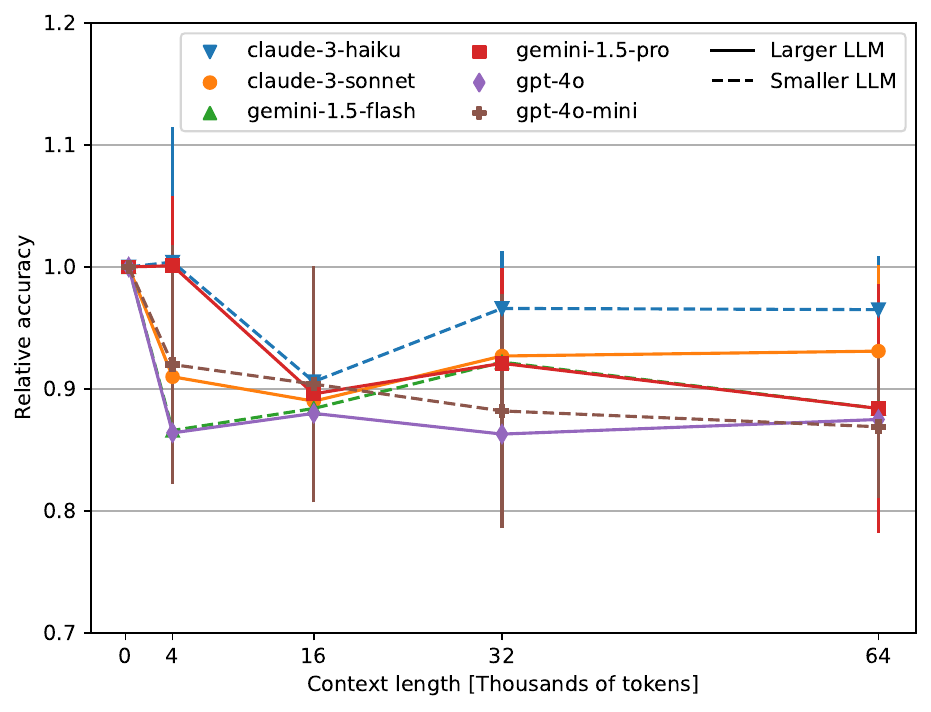}
    \caption{Experimental LLM performance, relative to the base \textit{no-context}  case, over variable \textit{free-chat} context. Scores are mean accuracies across three runs with the standard deviations shown as error bars. Dashed and continuous lines represent smaller and larger LLMs respectively.
}
    \label{fig:free_chat_relative}
\end{figure}

\begin{figure}[h!] 
    \centering
    \begin{subfigure}[b]{0.49\linewidth}
        \centering
            \includegraphics[width=\linewidth]{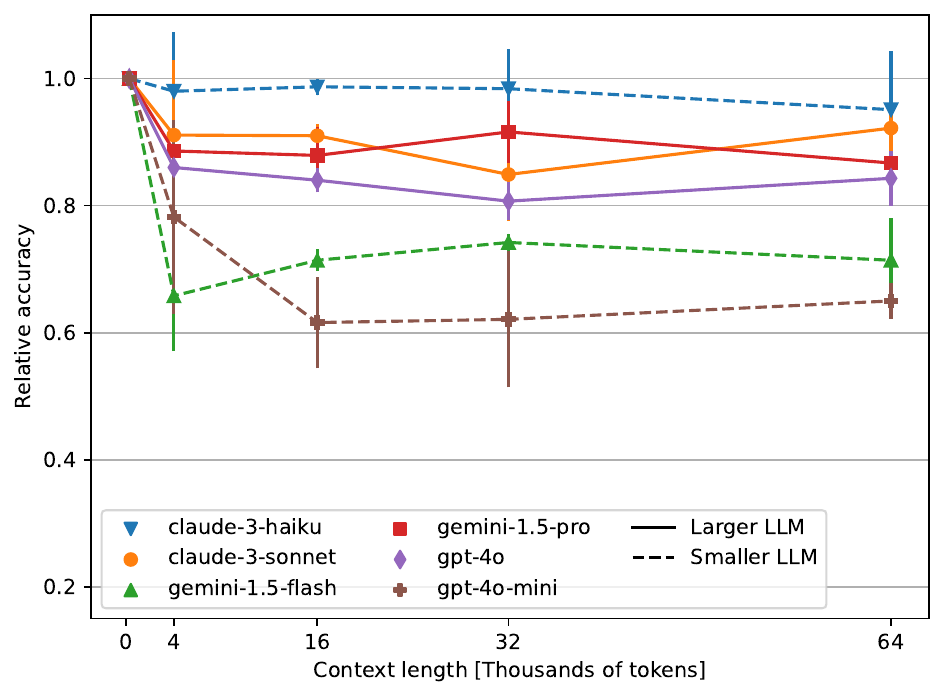}
    \caption{QA \textit{task-at-top} \textit{same-domain} context}
    \label{fig:same_domain_task_at_top_relative}
    \end{subfigure}
    \hfill
    \begin{subfigure}[b]{0.49\linewidth}
        \centering
        \includegraphics[width=\linewidth]{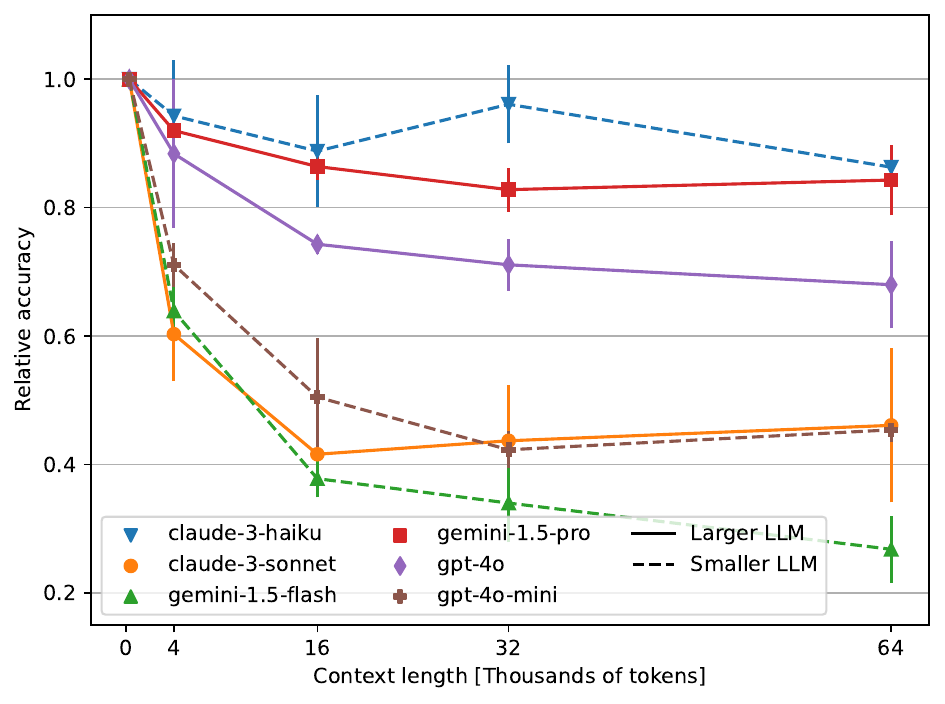}
    \caption{QA \textit{task-at-top} \textit{cross-domain} context}
    \label{fig:cross_domain_task_at_top_relative}
    \end{subfigure}
    \caption{Experimental LLM performance, relative to the base \textit{no-context}  case, over multi-turn \textit{task-at-top} \textit{same-domain} (a) and variable \textit{cross-domain} (b) context. Scores are mean relative accuracies across three runs with the standard deviations shown as error bars. Dashed and continuous lines represent smaller and larger LLMs respectively.}
    \label{fig:task_at_top_relative}
\end{figure}

\begin{figure}[h!] 
    \centering
    \begin{subfigure}[b]{0.49\linewidth}
        \centering
            \includegraphics[width=\linewidth]{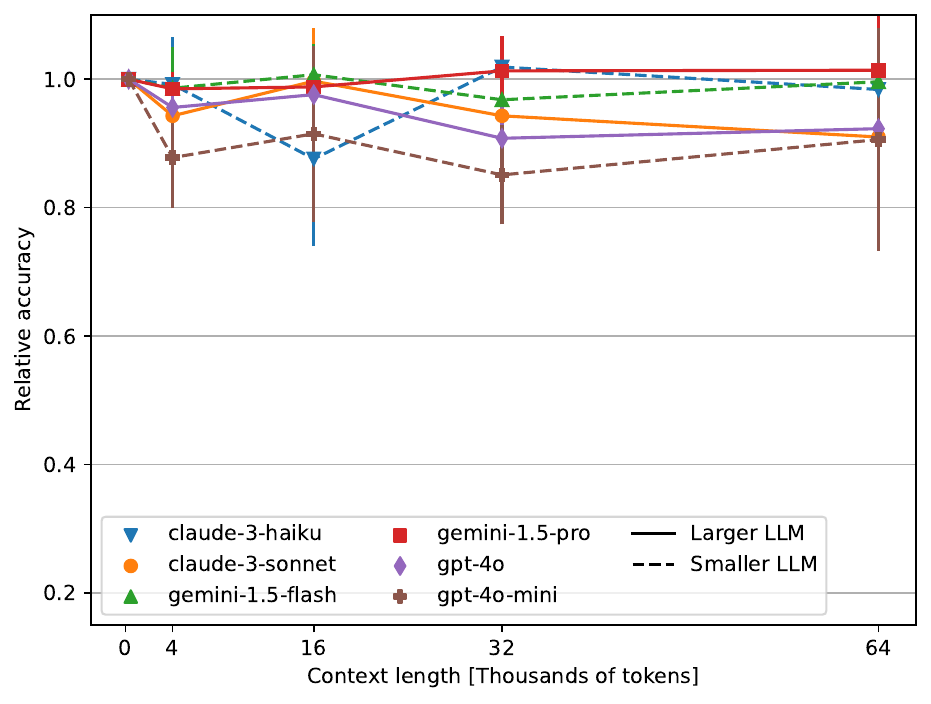}
    \caption{QA \textit{task-repeated} \textit{same-domain} context}
    \label{fig:same_domain_task_repeated_relative}
    \end{subfigure}
    \hfill
    \begin{subfigure}[b]{0.49\linewidth}
        \centering
        \includegraphics[width=\linewidth]{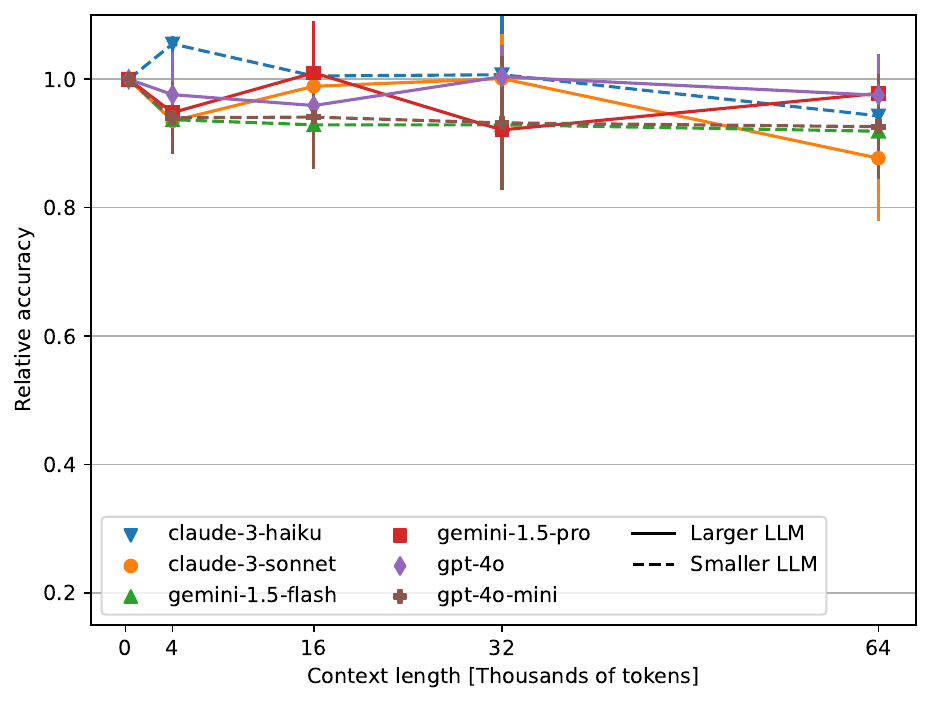}
    \caption{QA \textit{task-repeated} \textit{cross-domain} context}
    \label{fig:cross_domain_task_repeated_relative}
    \end{subfigure}
    \caption{Experimental LLM performance, relative to the base \textit{no-context}  case, over variable multi-turn \textit{task-repeated} \textit{same-domain} context (a) and \textit{cross-domain} (b) context. Scores are mean relative accuracies across three runs with the standard deviations shown as error bars. Dashed and continuous lines represent smaller and larger LLMs respectively.}
    \label{fig:task_repeated_relative}
\end{figure}

Figures \ref{fig:free_chat_relative}, \ref{fig:task_at_top_relative} and \ref{fig:task_repeated_relative} show relative accuracies for the main experiments with respect to the base \textit{no-context}  case.

\section{Task Location Experiment Results}

\begin{figure}[htbp] 
    \centering
    \begin{subfigure}[b]{0.49\linewidth}
        \centering
        \includegraphics[width=\linewidth]{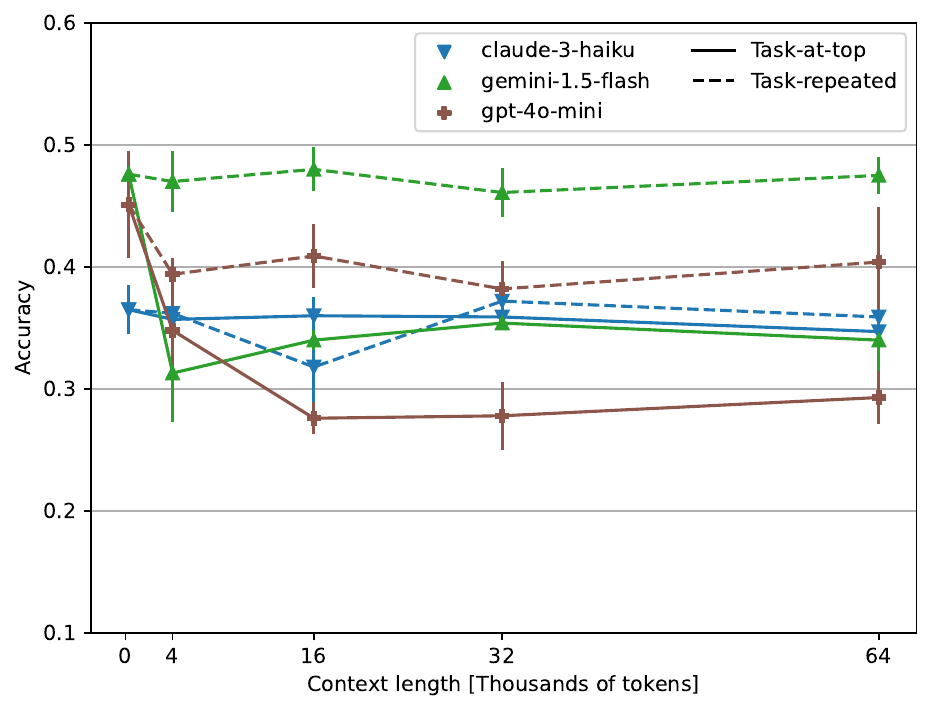}
        \caption{Smaller LLM}
        \label{fig:same - location - small}
    \end{subfigure}
    \hfill
    \begin{subfigure}[b]{0.49\linewidth}
        \centering
        \includegraphics[width=\linewidth]{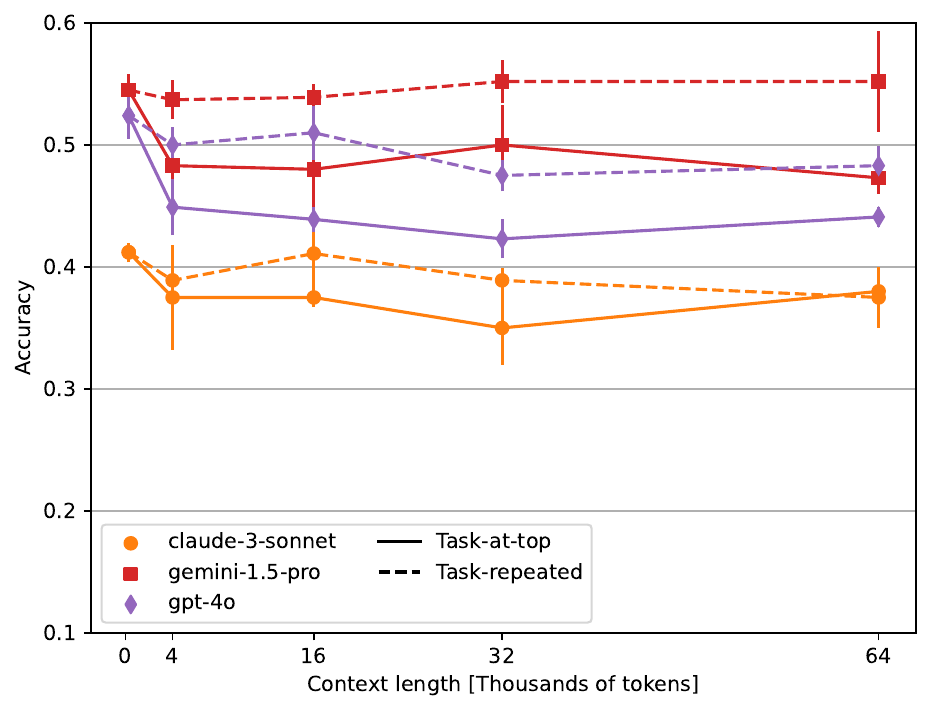}
        \caption{Larger LLM}
        \label{fig:same - location - large}
    \end{subfigure}
    \caption{Comparing smaller (a) and larger (b) LLM performances in \textit{task-at-top} and \textit{task-repeated} scenarios over \textit{same-domain} context. Scores are mean accuracies across three runs with the standard deviations shown as error bars. Continuous and dashed lines represent \textit{task-at-top} and \textit{task-repeated} experiments respectively.}
    \label{fig:same - location}
\end{figure}

\begin{figure}[htbp] 
    \centering
    \begin{subfigure}[b]{0.49\linewidth}
        \centering
        \includegraphics[width=\linewidth]{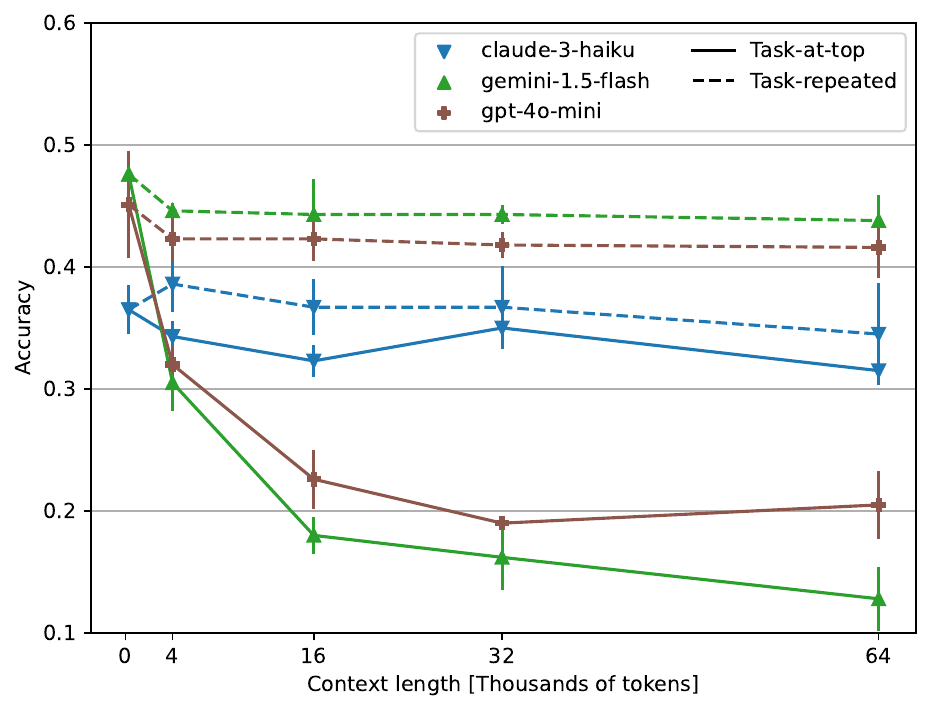}
        \caption{Smaller LLM}
        \label{fig:cross - location - small}
    \end{subfigure}
    \hfill
    \begin{subfigure}[b]{0.49\linewidth}
        \centering
        \includegraphics[width=\linewidth]{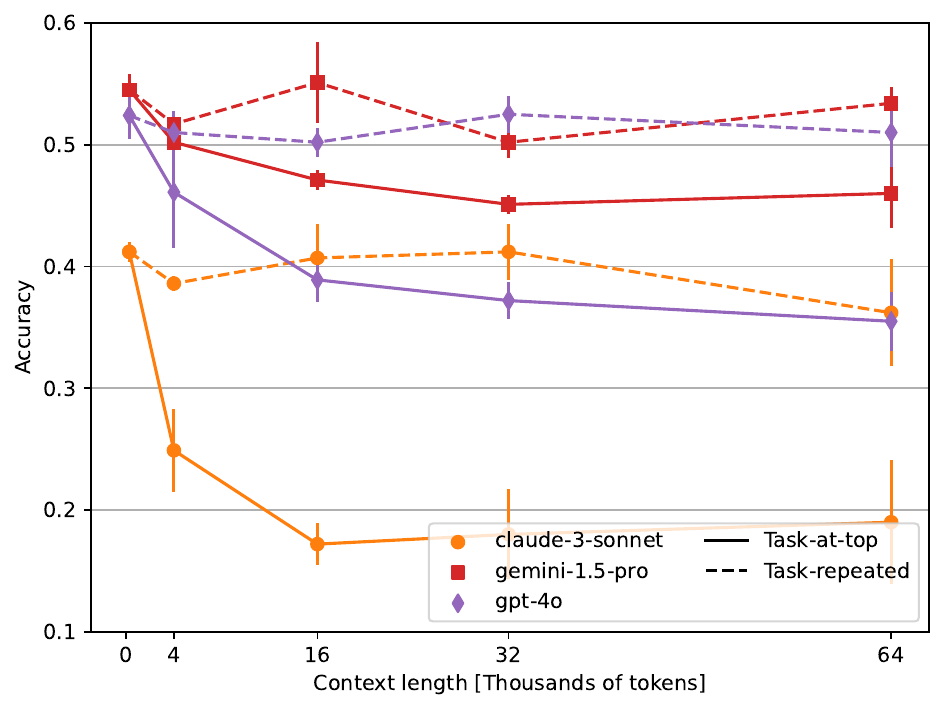}
        \caption{Larger LLM}
        \label{fig:cross - location - large}
    \end{subfigure}
    \caption{Comparing smaller (a) and larger (b) LLM performances in \textit{task-at-top} and \textit{task-repeated} scenarios over \textit{cross-domain} context. Scores are mean accuracies across three runs with the standard deviations shown as error bars. Continuous and dashed lines represent \textit{task-at-top} and \textit{task-repeated} experiments respectively.}
    \label{fig:cross - location}
\end{figure}

Figures \ref{fig:same - location} and  \ref{fig:cross - location} shows the comparison results for the task location experiments for the \textit{same-domain} and \textit{cross-domain} context.

\section{Unrelated-text Experiment}

We also explore the effect on model performance when Unrelated English text is added to the context prior to the target question. To do this we use sentences from the two most common books from Project Gutenburg: MobyDick \cite{melville1851} and Frankenstein \cite{shelley1818}. The unrelated text entails shuffled, concatenated passages of the books, chosen for their length and semantic distance from STEM domains, ensuring a substantial amount of unrelated context for the study. The different prior context lengths are extracted by hard truncation on the tokens level.
The task description for the target query is placed after the text blob and before the target query for the unrelated text context, as a separation between the two different content.\\

\begin{figure}[t]
    \centering
    \includegraphics[width=0.49\textwidth]{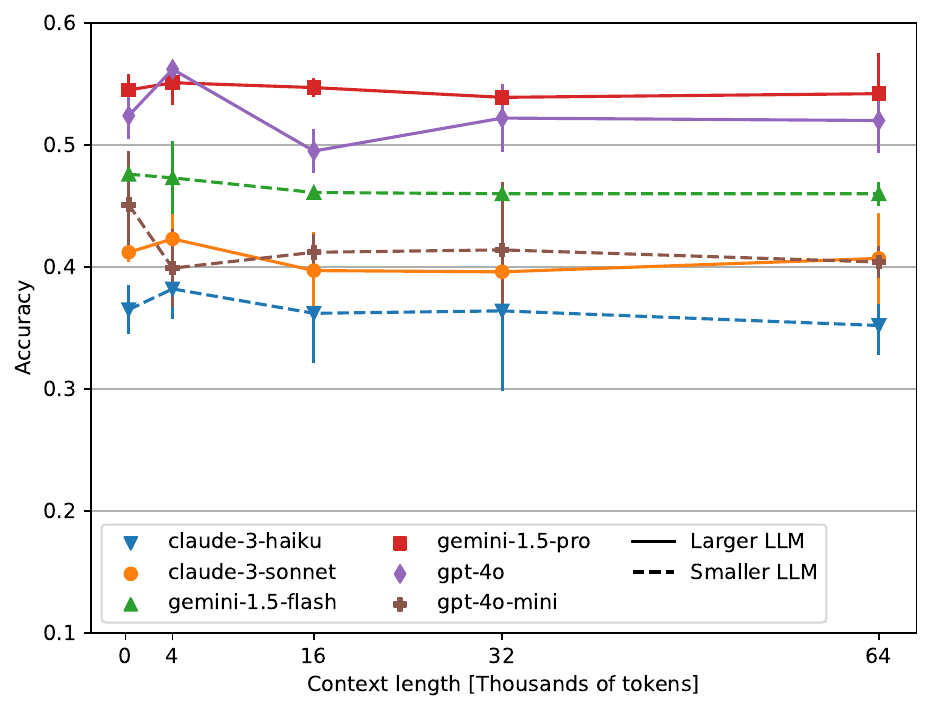}
    \caption{Experimental LLM performance over variable \textit{unrelated-text} context. Scores are mean accuracies across three runs with the standard deviations shown as error bars. Dashed and continuous lines represent smaller and larger LLMs respectively.
}
    \label{fig:unrelated_text}
\end{figure}

\noindent\textbf{Results.}
The presence of unrelated English text does not seem to significantly disrupt the performance of models as can bee seen in figure \ref{fig:unrelated_text}. 
Most models exhibit almost flat performance across various context lengths. 
The exception is GPT-4o-mini, where its accuracy drops by approximately 10\% in all experiments; however, the standard deviations of the accuracies are large, making it difficult to draw strong conclusions. 
We also observe many fluctuations, where sometimes the performance becomes slightly higher than in the baseline \textit{no-context}  scenario. 
This may be due to some slight model instability affecting the prediction probabilities, which can also be magnified by the non-zero value of the sampling temperature.

\section{LLM Models and Versions}
\label{appendix:models_versions}
Table \ref{table: models versions} shows what models and versions we used in our experiments.
\begin{table}[h]
    \centering
    \caption{LLM Models, Model version, and Services for text-completion APIs.}
      \begin{tabular}{lcc}
        \hline
        \textbf{Model} & \textbf{Model version} & \textbf{Service} \\
        \hline
        GPT-4o & \texttt{gpt-4o (2024-08-06)} & Azure OpenAI \\
        GPT-4o-mini & \texttt{gpt-4o mini (2024-07-18)} & Azure OpenAI \\
        Claude 3 Sonnet & \texttt{claude-3-sonnet-20240229-v1.0} & AWS Bedrock \\
        Claude 3 Haiku & \texttt{claude-3-haiku-20240307-v1.0} & AWS Bedrock \\
        Gemini 1.5 Pro & \texttt{gemini-1.5-pro-002} & GCP Vertex AI \\
        Gemini 1.5 Flash & \texttt{gemini-1.5-flash-002} & GCP Vertex AI \\
        \hline
      \end{tabular}
    \label{table: models versions}
  \end{table}

\newpage
\section{MMLU Subjects Selection}
\label{sec:appendix - mmlu}
We used the following STEM subject for the \textit{same-domain} context:
\begin{itemize}[noitemsep]
	\item High-School Biology
	\item High-School Physics
	\item High-School Chemistry
	\item Clinical Knowledge
	\item College Biology
	\item College Physics
	\item College Chemistry
	\item Anatomy
	\item Professional Medicine
	\item Virology
	\item Conceptual Physics
	\item Medical Genetics
	\item High-School Math
	\item College Math
	\item Abstract Algebra
\end{itemize}

We used the following Non-STEM subject for the \textit{cross-domain} context:
\begin{itemize}[noitemsep]
	\item Business
	\item Global fact
	\item High-School Geography
	\item High-School Government and Politics
	\item High-School Psychology
	\item High-School US History
	\item High-School World History
	\item International Law
	\item Jurisprudence
	\item Logical Fallacies
	\item Management
	\item Marketing
	\item Philosophy
	\item Professional Law
	\item Professional Psychology
	\item Public Relations
	\item Sociology
	\item US Foreign Policy
\end{itemize}

\section{Experiments Cost}
\label{appendix:experiments_cost}
Table \ref{table: models cost} shows the estimated cost of our experiments.
\begin{table}[h!]
    \centering
        \caption{\label{table: models cost}
              Experiment estimated cost from 15k API calls per model across 25 experiments and 3 runs.
        }
    \begin{tabular}{lr}
    \hline
    \textbf{Model}      & Cost (\$)   \\
    \hline
    Claude 3 Haiku      & 117 \\
    Claude 3 Sonnet     & 1406 \\
    Gemini 1.5 Flash    & 34 \\
    Gemini 1.5 Pro      & 573 \\
    GPT-4o              & 1167\\
    GPT-4o-mini         & 69 \\
    \hline
    \textbf{Total} & 3367 \\
    \end{tabular}
\end{table}

\section{LLM Usage}
\label{appendix:llm usage}
This paper was written by the authors, with GPT-4o used as a tool for refining grammar, spelling, and clarity.

\section{Data Usage}
\label{appendix:data}
This paper makes use of three datasets: LMSYS-1M-Chat \cite{zheng2024lmsyschat1mlargescalerealworldllm}, GPQA \cite{rein2023gpqa} and MMLU \cite{hendrycks2020measuring} datasets, all of which are available under open-source licenses. 
We adhered to their usage terms, ensuring proper citation and compliance with the license provisions. 
LMSYS-1M-Chat contains one million real-world conversations with state-of-the-art LLMs, curated for understanding and advancing LLM capabilities. 
GPQA and MMLU are MCQ-based datasets.
GPQA is an English-language dataset designed to evaluate general-purpose question-answering models, while MMLU is a large-scale benchmark that covers multiple subject areas to assess reasoning and knowledge. 
Additionally, we used Frankenstein \cite{shelley1818} and Moby-Dick \cite{melville1851}, both sourced from Project Gutenberg. 
These texts are in the public domain, and we used them in accordance with Project Gutenberg's terms. 
All assets were used appropriately within the scope of research and in compliance with their respective licenses.

\section{Full Results Table}
Table \ref{appendix: results - full} shows the full performance results for the main experiments, including mean accuracies and mean relative accuracies with respect to the base \textit{no-context}  experiment across three runs, and their standard deviation.

\newpage

\begin{longtable}{|c|c|c|c|}
    \caption{Full LLM performance results, in terms of accuracy, across different context scenarios (various sources and lengths). Accuracies are reported as mean values of three runs, with standard deviations.}
    \label{appendix: results - full} \\
    \hline
    \textbf{Model} & \textbf{Context Length} & \textbf{Accuracy} & \textbf{Relative Accuracy} \\
    \hline
    \endfirsthead
    \hline
    \textbf{Model} & \textbf{Context Length} & \textbf{Accuracy} & \textbf{Relative Accuracy} \\
    \hline
    \endhead
    \hline
    \endfoot
    \hline \hline
    \endlastfoot

Claude Haiku   &   no-context   &   0.365$\pm0.02$   &   1$\pm0$   \\
Claude Haiku   &   free-chat-4k   &   0.365$\pm0.021$   &   1.004$\pm0.111$   \\
Claude Haiku   &   unrelated-text-4k   &   0.382$\pm0.025$   &   1.046$\pm0.031$   \\
Claude Haiku   &   multi-turn-qa-4k-cross-domain-task-at-top   &   0.343$\pm0.013$   &   0.943$\pm0.087$   \\
Claude Haiku   &   multi-turn-qa-4k-cross-domain-task-repeated   &   0.386$\pm0.023$   &   1.055$\pm0.013$   \\
Claude Haiku   &   multi-turn-qa-4k-same-domain-task-at-top   &   0.357$\pm0.018$   &   0.98$\pm0.094$   \\
Claude Haiku   &   multi-turn-qa-4k-same-domain-task-repeated   &   0.362$\pm0.029$   &   0.992$\pm0.074$   \\
Claude Haiku   &   free-chat-16k   &   0.332$\pm0.036$   &   0.906$\pm0.048$   \\
Claude Haiku   &   unrelated-text-16k   &   0.362$\pm0.041$   &   0.99$\pm0.091$   \\
Claude Haiku   &   multi-turn-qa-16k-cross-domain-task-at-top   &   0.323$\pm0.013$   &   0.888$\pm0.087$   \\
Claude Haiku   &   multi-turn-qa-16k-cross-domain-task-repeated   &   0.367$\pm0.023$   &   1.005$\pm0.047$   \\
Claude Haiku   &   multi-turn-qa-16k-same-domain-task-at-top   &   0.36$\pm0.015$   &   0.987$\pm0.013$   \\
Claude Haiku   &   multi-turn-qa-16k-same-domain-task-repeated   &   0.318$\pm0.035$   &   0.876$\pm0.136$   \\
Claude Haiku   &   free-chat-32k   &   0.354$\pm0.036$   &   0.966$\pm0.047$   \\
Claude Haiku   &   unrelated-text-32k   &   0.364$\pm0.066$   &   0.991$\pm0.128$   \\
Claude Haiku   &   multi-turn-qa-32k-cross-domain-task-at-top   &   0.35$\pm0.008$   &   0.961$\pm0.061$   \\
Claude Haiku   &   multi-turn-qa-32k-cross-domain-task-repeated   &   0.367$\pm0.034$   &   1.007$\pm0.108$   \\
Claude Haiku   &   multi-turn-qa-32k-same-domain-task-at-top   &   0.359$\pm0.005$   &   0.984$\pm0.062$   \\
Claude Haiku   &   multi-turn-qa-32k-same-domain-task-repeated   &   0.372$\pm0.018$   &   1.019$\pm0.009$   \\
Claude Haiku   &   free-chat-64k   &   0.352$\pm0.006$   &   0.965$\pm0.044$   \\
Claude Haiku   &   unrelated-text-64k   &   0.352$\pm0.024$   &   0.963$\pm0.028$   \\
Claude Haiku   &   multi-turn-qa-64k-cross-domain-task-at-top   &   0.315$\pm0.006$   &   0.863$\pm0.034$   \\
Claude Haiku   &   multi-turn-qa-64k-cross-domain-task-repeated   &   0.345$\pm0.042$   &   0.943$\pm0.076$   \\
Claude Haiku   &   multi-turn-qa-64k-same-domain-task-at-top   &   0.347$\pm0.031$   &   0.951$\pm0.092$   \\
Claude Haiku   &   multi-turn-qa-64k-same-domain-task-repeated   &   0.359$\pm0.005$   &   0.984$\pm0.066$   \\
\hline
Claude Sonnet   &   no-context   &   0.412$\pm0.008$   &   1$\pm0$   \\
Claude Sonnet   &   free-chat-4k   &   0.375$\pm0.034$   &   0.91$\pm0.081$   \\
Claude Sonnet   &   unrelated-text-4k   &   0.423$\pm0.025$   &   1.025$\pm0.074$   \\
Claude Sonnet   &   multi-turn-qa-4k-cross-domain-task-at-top   &   0.249$\pm0.034$   &   0.603$\pm0.073$   \\
Claude Sonnet   &   multi-turn-qa-4k-cross-domain-task-repeated   &   0.386$\pm0.006$   &   0.935$\pm0.03$   \\
Claude Sonnet   &   multi-turn-qa-4k-same-domain-task-at-top   &   0.375$\pm0.043$   &   0.911$\pm0.118$   \\
Claude Sonnet   &   multi-turn-qa-4k-same-domain-task-repeated   &   0.389$\pm0.009$   &   0.943$\pm0.025$   \\
Claude Sonnet   &   free-chat-16k   &   0.367$\pm0.019$   &   0.89$\pm0.037$   \\
Claude Sonnet   &   unrelated-text-16k   &   0.397$\pm0.032$   &   0.963$\pm0.068$   \\
Claude Sonnet   &   multi-turn-qa-16k-cross-domain-task-at-top   &   0.172$\pm0.017$   &   0.416$\pm0.041$   \\
Claude Sonnet   &   multi-turn-qa-16k-cross-domain-task-repeated   &   0.407$\pm0.028$   &   0.989$\pm0.086$   \\
Claude Sonnet   &   multi-turn-qa-16k-same-domain-task-at-top   &   0.375$\pm0.008$   &   0.91$\pm0.018$   \\
Claude Sonnet   &   multi-turn-qa-16k-same-domain-task-repeated   &   0.411$\pm0.028$   &   0.997$\pm0.082$   \\
Claude Sonnet   &   free-chat-32k   &   0.382$\pm0.011$   &   0.927$\pm0.031$   \\
Claude Sonnet   &   unrelated-text-32k   &   0.396$\pm0.008$   &   0.959$\pm0.027$   \\
Claude Sonnet   &   multi-turn-qa-32k-cross-domain-task-at-top   &   0.18$\pm0.037$   &   0.437$\pm0.087$   \\
Claude Sonnet   &   multi-turn-qa-32k-cross-domain-task-repeated   &   0.412$\pm0.023$   &   1.001$\pm0.069$   \\
Claude Sonnet   &   multi-turn-qa-32k-same-domain-task-at-top   &   0.35$\pm0.03$   &   0.849$\pm0.073$   \\
Claude Sonnet   &   multi-turn-qa-32k-same-domain-task-repeated   &   0.389$\pm0.01$   &   0.943$\pm0.025$   \\
Claude Sonnet   &   free-chat-64k   &   0.384$\pm0.031$   &   0.931$\pm0.071$   \\
Claude Sonnet   &   unrelated-text-64k   &   0.407$\pm0.037$   &   0.987$\pm0.074$   \\
Claude Sonnet   &   multi-turn-qa-64k-cross-domain-task-at-top   &   0.19$\pm0.051$   &   0.461$\pm0.12$   \\
Claude Sonnet   &   multi-turn-qa-64k-cross-domain-task-repeated   &   0.362$\pm0.044$   &   0.877$\pm0.098$   \\
Claude Sonnet   &   multi-turn-qa-64k-same-domain-task-at-top   &   0.38$\pm0.016$   &   0.922$\pm0.037$   \\
Claude Sonnet   &   multi-turn-qa-64k-same-domain-task-repeated   &   0.375$\pm0.025$   &   0.91$\pm0.051$   \\
\hline
Gemini Flash   &   no-context   &   0.476$\pm0.006$   &   1$\pm0$   \\
Gemini Flash   &   free-chat-4k   &   0.412$\pm0.008$   &   0.866$\pm0.007$   \\
Gemini Flash   &   unrelated-text-4k   &   0.473$\pm0.03$   &   0.993$\pm0.071$   \\
Gemini Flash   &   multi-turn-qa-4k-cross-domain-task-at-top   &   0.305$\pm0.023$   &   0.639$\pm0.042$   \\
Gemini Flash   &   multi-turn-qa-4k-cross-domain-task-repeated   &   0.446$\pm0.006$   &   0.937$\pm0.021$   \\
Gemini Flash   &   multi-turn-qa-4k-same-domain-task-at-top   &   0.313$\pm0.04$   &   0.658$\pm0.087$   \\
Gemini Flash   &   multi-turn-qa-4k-same-domain-task-repeated   &   0.47$\pm0.025$   &   0.986$\pm0.064$   \\
Gemini Flash   &   free-chat-16k   &   0.421$\pm0.011$   &   0.884$\pm0.031$   \\
Gemini Flash   &   unrelated-text-16k   &   0.461$\pm0.003$   &   0.968$\pm0.01$   \\
Gemini Flash   &   multi-turn-qa-16k-cross-domain-task-at-top   &   0.18$\pm0.015$   &   0.378$\pm0.029$   \\
Gemini Flash   &   multi-turn-qa-16k-cross-domain-task-repeated   &   0.443$\pm0.029$   &   0.929$\pm0.053$   \\
Gemini Flash   &   multi-turn-qa-16k-diff-taskinlast   &   0.441$\pm0.024$   &   0.926$\pm0.054$   \\
Gemini Flash   &   multi-turn-qa-16k-same-domain-task-at-top   &   0.34$\pm0.008$   &   0.714$\pm0.017$   \\
Gemini Flash   &   multi-turn-qa-16k-same-domain-task-repeated   &   0.48$\pm0.018$   &   1.007$\pm0.048$   \\
Gemini Flash   &   free-chat-32k   &   0.439$\pm0.02$   &   0.922$\pm0.033$   \\
Gemini Flash   &   unrelated-text-32k   &   0.46$\pm0.005$   &   0.965$\pm0.022$   \\
Gemini Flash   &   multi-turn-qa-32k-cross-domain-task-at-top   &   0.162$\pm0.027$   &   0.34$\pm0.061$   \\
Gemini Flash   &   multi-turn-qa-32k-cross-domain-task-repeated   &   0.443$\pm0.008$   &   0.929$\pm0.007$   \\
Gemini Flash   &   multi-turn-qa-32k-same-domain-task-at-top   &   0.354$\pm0.005$   &   0.742$\pm0.014$   \\
Gemini Flash   &   multi-turn-qa-32k-same-domain-task-repeated   &   0.461$\pm0.02$   &   0.968$\pm0.038$   \\
Gemini Flash   &   free-chat-64k   &   0.421$\pm0.034$   &   0.884$\pm0.083$   \\
Gemini Flash   &   unrelated-text-64k   &   0.46$\pm0.01$   &   0.965$\pm0.032$   \\
Gemini Flash   &   multi-turn-qa-64k-cross-domain-task-at-top   &   0.128$\pm0.026$   &   0.268$\pm0.052$   \\
Gemini Flash   &   multi-turn-qa-64k-cross-domain-task-repeated   &   0.438$\pm0.021$   &   0.919$\pm0.048$   \\
Gemini Flash   &   multi-turn-qa-64k-same-domain-task-at-top   &   0.34$\pm0.028$   &   0.714$\pm0.067$   \\
Gemini Flash   &   multi-turn-qa-64k-same-domain-task-repeated   &   0.475$\pm0.015$   &   0.996$\pm0.022$   \\
\hline
Gemini Pro   &   no-context   &   0.545$\pm0.013$   &   1$\pm0$   \\
Gemini Pro   &   free-chat-4k   &   0.545$\pm0.018$   &   1.001$\pm0.057$   \\
Gemini Pro   &   unrelated-text-4k   &   0.551$\pm0.018$   &   1.009$\pm0.027$   \\
Gemini Pro   &   multi-turn-qa-4k-cross-domain-task-at-top   &   0.502$\pm0.008$   &   0.92$\pm0.013$   \\
Gemini Pro   &   multi-turn-qa-4k-cross-domain-task-repeated   &   0.517$\pm0.011$   &   0.948$\pm0.032$   \\
Gemini Pro   &   multi-turn-qa-4k-same-domain-task-at-top   &   0.483$\pm0.018$   &   0.886$\pm0.02$   \\
Gemini Pro   &   multi-turn-qa-4k-same-domain-task-repeated   &   0.537$\pm0.016$   &   0.985$\pm0.026$   \\
Gemini Pro   &   free-chat-16k   &   0.488$\pm0.015$   &   0.896$\pm0.049$   \\
Gemini Pro   &   unrelated-text-16k   &   0.547$\pm0.008$   &   1.003$\pm0.024$   \\
Gemini Pro   &   multi-turn-qa-16k-cross-domain-task-at-top   &   0.471$\pm0.008$   &   0.864$\pm0.021$   \\
Gemini Pro   &   multi-turn-qa-16k-cross-domain-task-repeated   &   0.551$\pm0.033$   &   1.01$\pm0.08$   \\
Gemini Pro   &   multi-turn-qa-16k-same-domain-task-at-top   &   0.48$\pm0.033$   &   0.879$\pm0.045$   \\
Gemini Pro   &   multi-turn-qa-16k-same-domain-task-repeated   &   0.539$\pm0.011$   &   0.988$\pm0.021$   \\
Gemini Pro   &   free-chat-32k   &   0.502$\pm0.03$   &   0.921$\pm0.078$   \\
Gemini Pro   &   unrelated-text-32k   &   0.539$\pm0.006$   &   0.988$\pm0.019$   \\
Gemini Pro   &   multi-turn-qa-32k-cross-domain-task-at-top   &   0.451$\pm0.008$   &   0.828$\pm0.034$   \\
Gemini Pro   &   multi-turn-qa-32k-cross-domain-task-repeated   &   0.502$\pm0.013$   &   0.921$\pm0.046$   \\
Gemini Pro   &   multi-turn-qa-32k-same-domain-task-at-top   &   0.5$\pm0.033$   &   0.916$\pm0.049$   \\
Gemini Pro   &   multi-turn-qa-32k-same-domain-task-repeated   &   0.552$\pm0.018$   &   1.013$\pm0.055$   \\
Gemini Pro   &   free-chat-64k   &   0.481$\pm0.044$   &   0.884$\pm0.102$   \\
Gemini Pro   &   unrelated-text-64k   &   0.542$\pm0.033$   &   0.995$\pm0.082$   \\
Gemini Pro   &   multi-turn-qa-64k-cross-domain-task-at-top   &   0.46$\pm0.028$   &   0.843$\pm0.055$   \\
Gemini Pro   &   multi-turn-qa-64k-cross-domain-task-repeated   &   0.534$\pm0.013$   &   0.978$\pm0.005$   \\
Gemini Pro   &   multi-turn-qa-64k-same-domain-task-at-top   &   0.473$\pm0.013$   &   0.867$\pm0.013$   \\
Gemini Pro   &   multi-turn-qa-64k-same-domain-task-repeated   &   0.552$\pm0.041$   &   1.014$\pm0.095$   \\
\hline
GPT-4o   &   no-context   &   0.524$\pm0.019$   &   1$\pm0$   \\
GPT-4o   &   free-chat-4k   &   0.453$\pm0.03$   &   0.864$\pm0.029$   \\
GPT-4o   &   unrelated-text-4k   &   0.562$\pm0.003$   &   1.075$\pm0.042$   \\
GPT-4o   &   multi-turn-qa-4k-cross-domain-task-at-top   &   0.461$\pm0.046$   &   0.884$\pm0.116$   \\
GPT-4o   &   multi-turn-qa-4k-cross-domain-task-repeated   &   0.51$\pm0.018$   &   0.976$\pm0.067$   \\
GPT-4o   &   multi-turn-qa-4k-same-domain-task-at-top   &   0.449$\pm0.023$   &   0.86$\pm0.062$   \\
GPT-4o   &   multi-turn-qa-4k-same-domain-task-repeated   &   0.5$\pm0.015$   &   0.956$\pm0.048$   \\
GPT-4o   &   free-chat-16k   &   0.461$\pm0.031$   &   0.88$\pm0.028$   \\
GPT-4o   &   unrelated-text-16k   &   0.495$\pm0.018$   &   0.947$\pm0.065$   \\
GPT-4o   &   multi-turn-qa-16k-cross-domain-task-at-top   &   0.389$\pm0.018$   &   0.743$\pm0.015$   \\
GPT-4o   &   multi-turn-qa-16k-cross-domain-task-repeated   &   0.502$\pm0.012$   &   0.959$\pm0.047$   \\
GPT-4o   &   multi-turn-qa-16k-same-domain-task-at-top   &   0.439$\pm0.01$   &   0.84$\pm0.019$   \\
GPT-4o   &   multi-turn-qa-16k-same-domain-task-repeated   &   0.51$\pm0.022$   &   0.976$\pm0.066$   \\
GPT-4o   &   free-chat-32k   &   0.451$\pm0.013$   &   0.863$\pm0.051$   \\
GPT-4o   &   unrelated-text-32k   &   0.522$\pm0.028$   &   0.998$\pm0.062$   \\
GPT-4o   &   multi-turn-qa-32k-cross-domain-task-at-top   &   0.372$\pm0.015$   &   0.711$\pm0.04$   \\
GPT-4o   &   multi-turn-qa-32k-cross-domain-task-repeated   &   0.525$\pm0.015$   &   1.004$\pm0.05$   \\
GPT-4o   &   multi-turn-qa-32k-same-domain-task-at-top   &   0.423$\pm0.016$   &   0.807$\pm0.03$   \\
GPT-4o   &   multi-turn-qa-32k-same-domain-task-repeated   &   0.475$\pm0.013$   &   0.908$\pm0.057$   \\
GPT-4o   &   free-chat-64k   &   0.458$\pm0.013$   &   0.875$\pm0.035$   \\
GPT-4o   &   unrelated-text-64k   &   0.52$\pm0.027$   &   0.996$\pm0.085$   \\
GPT-4o   &   multi-turn-qa-64k-cross-domain-task-at-top   &   0.355$\pm0.024$   &   0.68$\pm0.068$   \\
GPT-4o   &   multi-turn-qa-64k-cross-domain-task-repeated   &   0.51$\pm0.028$   &   0.975$\pm0.065$   \\
GPT-4o   &   multi-turn-qa-64k-same-domain-task-at-top   &   0.441$\pm0.008$   &   0.843$\pm0.043$   \\
GPT-4o   &   multi-turn-qa-64k-same-domain-task-repeated   &   0.483$\pm0.016$   &   0.923$\pm0.024$   \\
\hline
GPT-4o-mini   &   no-context   &   0.451$\pm0.044$   &   1$\pm0$   \\
GPT-4o-mini   &   free-chat-4k   &   0.412$\pm0.011$   &   0.92$\pm0.098$   \\
GPT-4o-mini   &   unrelated-text-4k   &   0.399$\pm0.032$   &   0.892$\pm0.127$   \\
GPT-4o-mini   &   multi-turn-qa-4k-cross-domain-task-at-top   &   0.32$\pm0.016$   &   0.711$\pm0.034$   \\
GPT-4o-mini   &   multi-turn-qa-4k-cross-domain-task-repeated   &   0.423$\pm0.019$   &   0.94$\pm0.056$   \\
GPT-4o-mini   &   multi-turn-qa-4k-diff-taskinlast   &   0.402$\pm0.028$   &   0.901$\pm0.145$   \\
GPT-4o-mini   &   multi-turn-qa-4k-same-domain-task-at-top   &   0.348$\pm0.036$   &   0.782$\pm0.152$   \\
GPT-4o-mini   &   multi-turn-qa-4k-same-domain-task-repeated   &   0.394$\pm0.013$   &   0.878$\pm0.079$   \\
GPT-4o-mini   &   free-chat-16k   &   0.406$\pm0.026$   &   0.904$\pm0.097$   \\
GPT-4o-mini   &   unrelated-text-16k   &   0.412$\pm0.015$   &   0.918$\pm0.054$   \\
GPT-4o-mini   &   multi-turn-qa-16k-cross-domain-task-at-top   &   0.226$\pm0.024$   &   0.505$\pm0.092$   \\
GPT-4o-mini   &   multi-turn-qa-16k-cross-domain-task-repeated   &   0.423$\pm0.018$   &   0.941$\pm0.081$   \\
GPT-4o-mini   &   multi-turn-qa-16k-same-domain-task-at-top   &   0.276$\pm0.013$   &   0.616$\pm0.072$   \\
GPT-4o-mini   &   multi-turn-qa-16k-same-domain-task-repeated   &   0.409$\pm0.026$   &   0.915$\pm0.137$   \\
GPT-4o-mini   &   free-chat-32k   &   0.396$\pm0.026$   &   0.882$\pm0.096$   \\
GPT-4o-mini   &   unrelated-text-32k   &   0.414$\pm0.056$   &   0.92$\pm0.119$   \\
GPT-4o-mini   &   multi-turn-qa-32k-cross-domain-task-at-top   &   0.19$\pm0.006$   &   0.423$\pm0.029$   \\
GPT-4o-mini   &   multi-turn-qa-32k-cross-domain-task-repeated   &   0.418$\pm0.011$   &   0.932$\pm0.104$   \\
GPT-4o-mini   &   multi-turn-qa-32k-same-domain-task-at-top   &   0.278$\pm0.028$   &   0.621$\pm0.107$   \\
GPT-4o-mini   &   multi-turn-qa-32k-same-domain-task-repeated   &   0.382$\pm0.023$   &   0.851$\pm0.077$   \\
GPT-4o-mini   &   free-chat-64k   &   0.392$\pm0.048$   &   0.869$\pm0.058$   \\
GPT-4o-mini   &   unrelated-text-64k   &   0.404$\pm0.013$   &   0.901$\pm0.089$   \\
GPT-4o-mini   &   multi-turn-qa-64k-cross-domain-task-at-top   &   0.205$\pm0.028$   &   0.454$\pm0.019$   \\
GPT-4o-mini   &   multi-turn-qa-64k-cross-domain-task-repeated   &   0.416$\pm0.025$   &   0.926$\pm0.082$   \\
GPT-4o-mini   &   multi-turn-qa-64k-same-domain-task-at-top   &   0.293$\pm0.022$   &   0.65$\pm0.028$   \\
  
\end{longtable}

\end{document}